\documentclass{ar-1col-S2O}
\usepackage[numbers]{natbib}
\usepackage{url}
\usepackage{hyperref}
\usepackage{amsmath}
\usepackage{amssymb}
\jname{Annu. Rev. Control Robot. Auton. Syst.}
\jvol{AA}
\jyear{YYYY}
\doi{10.1146/((please add article doi))}

\begin{document}

\markboth{Hong et al.}{Control of Marine Robots in the Era of Data-Driven Intelligence}

\title{Control of Marine Robots in the Era of Data-Driven Intelligence}

\author{Lin Hong,$^{1,2}$ Lu Liu,$^3$ Zhouhua Peng,$^3$ and Fumin Zhang$^{1,2,*}$
\affil{$^1$Department of Electronic and Computer Engineering, The Hong Kong University of Science and Technology, Hong Kong, China; email:eelinhong@ust.hk,eefumin@ust.hk}
\affil{$^2$Cheng-Kar Shun Robotics Institute, The Hong Kong University of Science and Technology, Hong Kong, China.}
\affil{$^3$School of Marine Electrical Engineering, Dalian Maritime University, Dalian, China; email: luliu@dlmu.edu.cn, zhpeng@dlmu.edu.cn}
\affil{$^*$ Corresponding author}
}

\begin{abstract}
The control of marine robots has long relied on model-based methods grounded in classical and modern control theory. However, the nonlinearity and uncertainties inherent in robot dynamics, coupled with the complexity of marine environments, have revealed the limitations of conventional control methods. 
The rapid evolution of machine learning have opened new avenues for incorporating data-driven intelligence into control strategies, prompting a paradigm shift in the control of marine robots.
This paper provides a review of recent progress in marine robot control through the lens of this emerging paradigm. The review covers both individual and cooperative marine robotic systems, highlighting notable achievements in data-driven control of marine robots and summarizing open-source resources that support the development and validation of advanced control methods.
Finally, several future perspectives are outlined to guide research toward achieving high-level autonomy for marine robots in real-world applications.
This paper aims to serve as a roadmap toward the next-generation control framework of marine robots in the era of data-driven intelligence.
\end{abstract}

\begin{keywords}
marine robots, model-based control, data-driven control, cooperative control, reinforcement learning, open source
\end{keywords}
\maketitle


\section{INTRODUCTION}
\begin{figure}[h]
\includegraphics[width=6.3in]{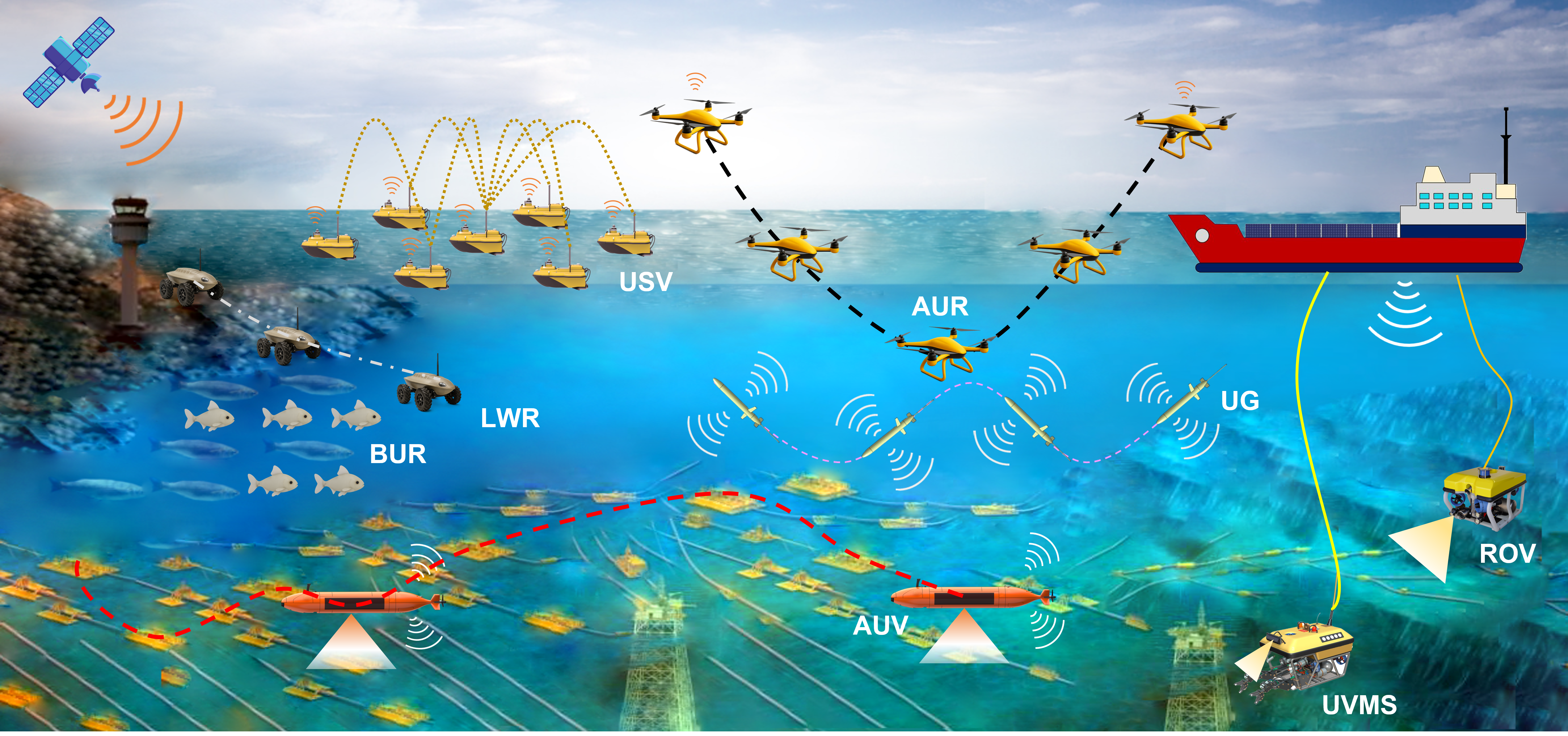}
\caption{Configurations of different types of marine robots and their practical applications in marine environments.}
\label{robots}
\end{figure}

\begin{figure}[h]
\includegraphics[width=3.7in]{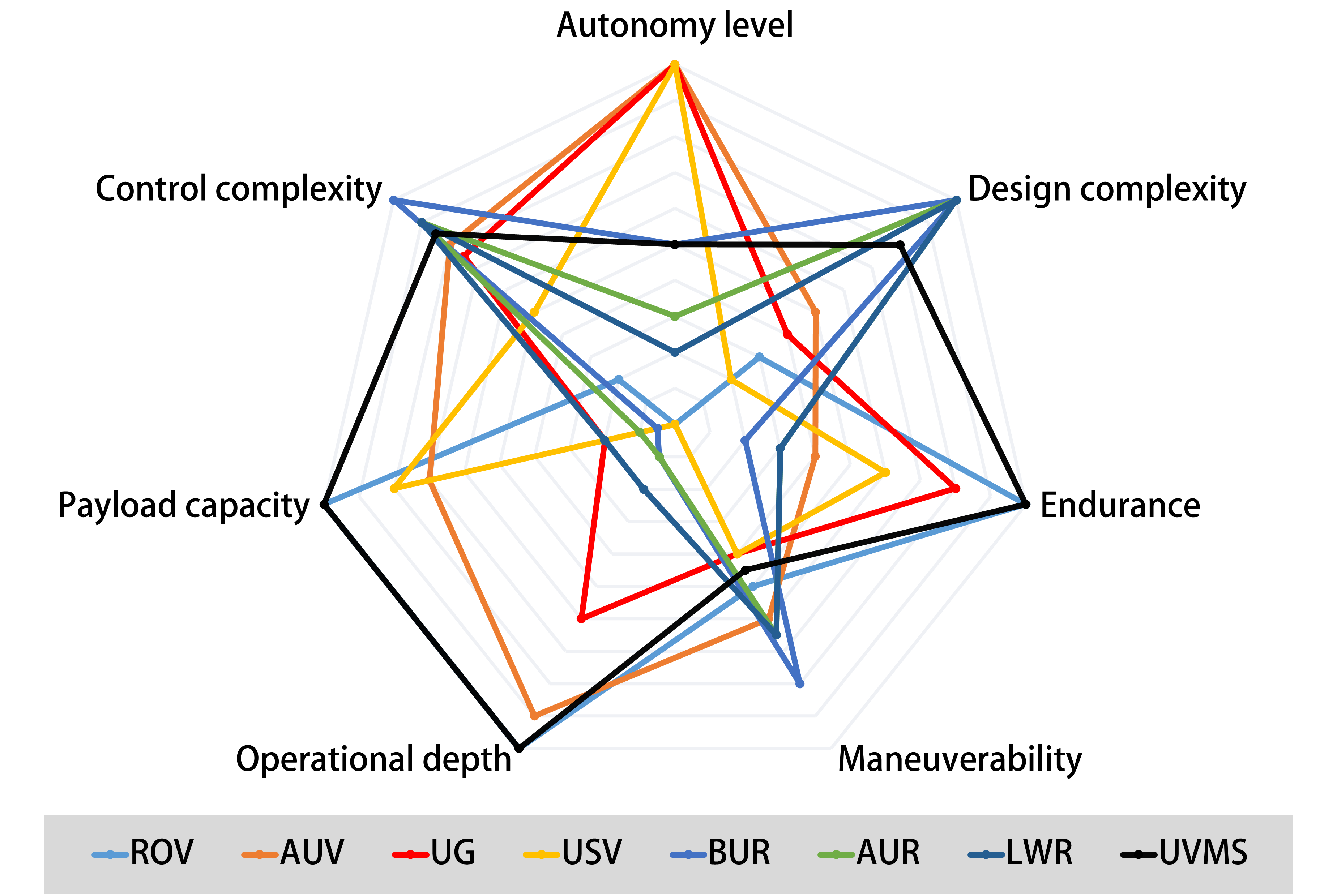}
\caption{Comparison of different types of marine robots in terms of autonomy level, design complexity, endurance, maneuverability, operational depth, payload capacity, and control complexity.}
\label{Robotcompare}
\end{figure}

Marine robots have been highly instrumental in unveiling the ocean’s mysteries since their emergence in the early 1970s~\cite{zhang2015future}. 
Over the past few decades, marine robots have diversified into specialized types, including remotely operated vehicles (ROVs), autonomous underwater vehicles (AUVs), unmanned surface vehicles (USVs), underwater gliders (UGs), biomimetic underwater robots (BURs), aerial-underwater robots (AURs), land-water robots (LWRs), and underwater vehicle-manipulator systems (UVMSs) (\textbf{See Figure~\ref{robots}}). Each type is designed with unique features (\textbf{See Figure~\ref{Robotcompare}}) to meet growing demands in tasks such as offshore infrastructure inspection~\cite{campos2024nautilus}, marine life observation~\cite{katzschmann2018exploration}, autonomous underwater manipulation~\cite{wang2023versatile}, and cross-domain data collection~\cite{10814088}.
With the increasing scale and complexity of underwater tasks (e.g., ocean sampling~\cite{leonard2010coordinated}, seabed mapping~\cite{paull2018probabilistic}, etc.), single marine robots face limitations in sensing, communication, and computation.
To overcome these limitations, the development of cooperative multiple marine robotic systems has become a key technological frontier. In such systems, the collective behavior of multiple marine robots offers distinct advantages over single sophisticated individuals, including broader spatial coverage, enhanced robustness, greater scalability, and improved efficiency~\cite{wang2017cooperative}.

\begin{marginnote}[]
\entry{Mrine robots}{Members of the marine robot family discussed in the paper include ROVs, AUVs, USVs, UGs, BURs, AURs, LWRs, and UVMSs}
\end{marginnote}

Regardless of physical configuration or organizational structure, control systems are fundamental to marine robots, shaping their autonomy, stability, and reliability. Unlike terrestrial or aerial robots, marine robots operate in dynamic, unstructured fluid environments, facing nonlinear hydrodynamic forces and disturbances such as wind, currents, and waves. Their control systems must ensure capabilities like station-keeping, path following, and trajectory tracking, while remaining robust and adaptive to both system dynamics and environmental disturbances. Beyond effective control of individual marine robots, it is also essential to enable coordination among multiple robots, whether organized into structured teams or decentralized swarms, to perform complex tasks like flow fields mapping~\cite{chang2017motion} and cooperative underwater manipulation~\cite{Heshmati-JOE-2020}.

Control, often referred to as the cerebellum of marine robots, remains a vigorous research field. Prior reviews target specific platforms such as AUVs~\cite{degorre2023survey}, ROVs~\cite{he2020review}, USVs~\cite{wang2019state,er2023intelligent}, UGs~\cite{wang2022development}, and BURs~\cite{wang2020development,sun2022recent}, or focus on particular methods like fuzzy logic control~\cite{xiang2018survey}, MPC~\cite{wei2022mpc}, and data-driven control~\cite{hassani2018data}. More recent work surveys AUV controllers with experimental comparisons~\cite{tijjani2022survey} and underscores the rising role of single- and multi-agent autonomy, spurring advances in coordinated control~\cite{zhang2015future,das2016cooperative,peng2020overview,yang2021survey}. Yet these reviews are still confined to certain robot types or traditional model-based strategies, underscoring the need for a comprehensive survey of data-driven control spanning both single- and multi-robot marine systems.
To address this need, this paper presents a comprehensive review of recent advances in data-driven control for marine robots, encompassing both individual and cooperative systems. It also summarizes open-source resources, including simulation platforms and marine robotic systems, that support the development and validation of advanced control strategies. Lastly, several future research perspectives are outlined.

\section{COMMON MATHEMATICAL MODEL}
\label{sec:MMMR}
For rigid-body marine robots operating in a 3D workspace with six degrees of freedom (DoFs), their kinematics and dynamics can be described using a compact vectorial form. Consider a fleet of $N$ marine robots, each indexed by $i = 1, 2, \dots, N$.
The pose of the $i$-th marine robot in the inertial (earth-fixed) frame is defined as:
\begin{equation}
    \boldsymbol{\eta}_i = 
    \begin{bmatrix}
        \boldsymbol{\eta}_{i,1}^\mathrm{T} & \boldsymbol{\eta}_{i,2}^\mathrm{T}
    \end{bmatrix}^\mathrm{T}
    = 
    \begin{bmatrix}
        x_i & y_i & z_i & \phi_i & \theta_i & \psi_i
    \end{bmatrix}^\mathrm{T}
    \in \mathbb{R}^{6}
\end{equation}
where $\boldsymbol{\eta}_{i,1} = [\,x_i,\;y_i,\;z_i\,]^{\mathrm T} \in \mathbb{R}^{3}$ denotes the position and $\boldsymbol{\eta}_{i,2} = [\,\phi_i,\;\theta_i,\;\psi_i\,]^{\mathrm T} \in \mathbb{R}^{3}$ represents the Euler angles.

The body-fixed velocity vector is:
\begin{equation}
    \boldsymbol{v}_i = 
    \begin{bmatrix}
        \boldsymbol{v}_{i,1}^\mathrm{T} & \boldsymbol{v}_{i,2}^\mathrm{T}
    \end{bmatrix}^\mathrm{T}
    = 
    \begin{bmatrix}
        u_i & v_i & w_i & p_i & q_i & r_i
    \end{bmatrix}^\mathrm{T}
    \in \mathbb{R}^{6}
\end{equation}
where $\boldsymbol{v}_{i,1}= [\,u_i,\;v_i,\;w_i\,]^\mathrm{T}  \in \mathbb{R}^{3}$ and $\boldsymbol{v}_{i,2} = [\,p_i,\;q_i,\;r_i\,]^\mathrm{T} \in \mathbb{R}^{3}$ are the linear and angular velocity components, respectively.

The kinematic equation relates the time derivative of the pose to the velocity:
\begin{equation}
    \dot{\boldsymbol{\eta}}_i = \boldsymbol{J}(\boldsymbol{\eta}_{i,2}) \boldsymbol{v}_i
\end{equation}
where $\boldsymbol{J}(\boldsymbol{\eta}_{i,2})$ is the Jacobian transformation matrix:
\begin{equation}
    \boldsymbol{J}(\boldsymbol{\eta}_{i,2}) = 
    \begin{bmatrix}
        \boldsymbol{R}(\boldsymbol{\eta}_{i,2}) & \boldsymbol{0}_{3 \times 3} \\
        \boldsymbol{0}_{3 \times 3} & \boldsymbol{T}(\boldsymbol{\eta}_{i,2})
    \end{bmatrix}
\end{equation}
with $\boldsymbol{R}$ the rotation matrix from body to inertial frame and $\boldsymbol{T}$ the transformation matrix from angular velocity to Euler angle rates.

The dynamic model of the $i$-th robot can be described as:
\begin{equation}
    \boldsymbol{M}_i \dot{\boldsymbol{v}}_i + \boldsymbol{C}_i(\boldsymbol{v}_i) \boldsymbol{v}_i + \boldsymbol{D}_i(\boldsymbol{v}_i) \boldsymbol{v}_i + \boldsymbol{g}_i(\boldsymbol{\eta}_i) = \boldsymbol{\tau}_i + \boldsymbol{\tau}_{\text{env}} 
    \label{dm}
\end{equation}
where $\boldsymbol{M}_i$ is the inertia matrix (including added mass), $\boldsymbol{C}_i(\boldsymbol{v}_i)$ is the Coriolis and centripetal matrix, $\boldsymbol{D}_i(\boldsymbol{v}_i)$ is the hydrodynamic damping matrix, $\boldsymbol{g}_i(\boldsymbol{\eta}_i)$ represents the hydrostatic restoring forces and moments, $\boldsymbol{\tau}_i$ denotes the control input vector (forces and torques), and $\boldsymbol{\tau}_{\text{env}}$ accounts for environmental disturbances.

\begin{summary}[SUMMARY POINTS]
The above formulation provides a unified framework for controlling both single ($i=1$) and multiple ($i\geq2$) marine robots in real-world marine environments.
Based on Equation~\ref{dm} and the preceding analysis, several key challenges in marine robot control are identified:
\emph{1}) \textbf{Inherent nonlinearity}: Marine robots exhibit strong nonlinear dynamics due to fluid–structure interactions (FSI), limiting the effectiveness of linear control methods.
\emph{2}) \textbf{Model uncertainties}: Accurate modeling of actuation mechanisms (e.g., thrusters, fins) and hydrodynamic parameters (e.g., added mass, damping) is difficult, leading to model uncertainties.
\emph{3}) \textbf{External disturbances}: Environmental forces such as currents, waves, and wind can destabilize the system and degrade control performance.
\emph{4}) \textbf{Underactuation and input saturation}: Structural constraints often result in underactuated systems, and actuator limitations introduce input saturation, introducing physical constraints in control method design.
\emph{5}) \textbf{Unmeasured states}: Due to size and sensor limitations, marine robots frequently operate with partial state observations, requiring robust estimation or observer-based control strategies.
For multi-robot systems, an additional challenge is:
\emph{6}) \textbf{Communication constraints}: Limited bandwidth, high latency, and unreliable links in underwater environments hinder timely information sharing and coordination among marine robots.
\end{summary}

\section{CONTROL OF SINGLE MARINE ROBOT} 
The control of a single marine robot has increasingly shifted from conventional methods toward data-driven methods, driven by the rapid advances and proven capabilities of machine learning.
This section reviews recent progress in data-driven control strategies, which are classified into three categories: \emph{1)} model-based data-driven methods, \emph{2)} model-free data-driven methods, and \emph{3)} hybrid approaches that integrate conventional control methods and data-driven algorithms. A summary of model-based control methods for marine robots, validated through practical experiments, is provided in \textbf{Supplemental Material 1}.

\begin{marginnote}[]
\entry{\href{https://zenodo.org/records/15730809?token=eyJhbGciOiJIUzUxMiJ9.eyJpZCI6ImQ5MDI4ODUzLTk2YzEtNDcxMi1iNTdiLTg4MTFkMDQ3NmYzOCIsImRhdGEiOnt9LCJyYW5kb20iOiJmMWQ4NzdhNWYwMTc3YzQ5NDQyMTE5ODZhZWMxY2U5MyJ9.nYXNZs5kJQ9yGZ0vQn1YWuXn9oWmSPRN6l5-_mtn0p8LvMv8D1WuyQnrv2NOjp5ozhnYmsm59iBkiZUObA-7SA}{Supplemental Material 1}}{A summary of model-based control methods for marine robots.}
\end{marginnote}

\subsection{Model-Based Data-Driven Control Methods}
\begin{figure}[h]
\includegraphics[width=5.5in]{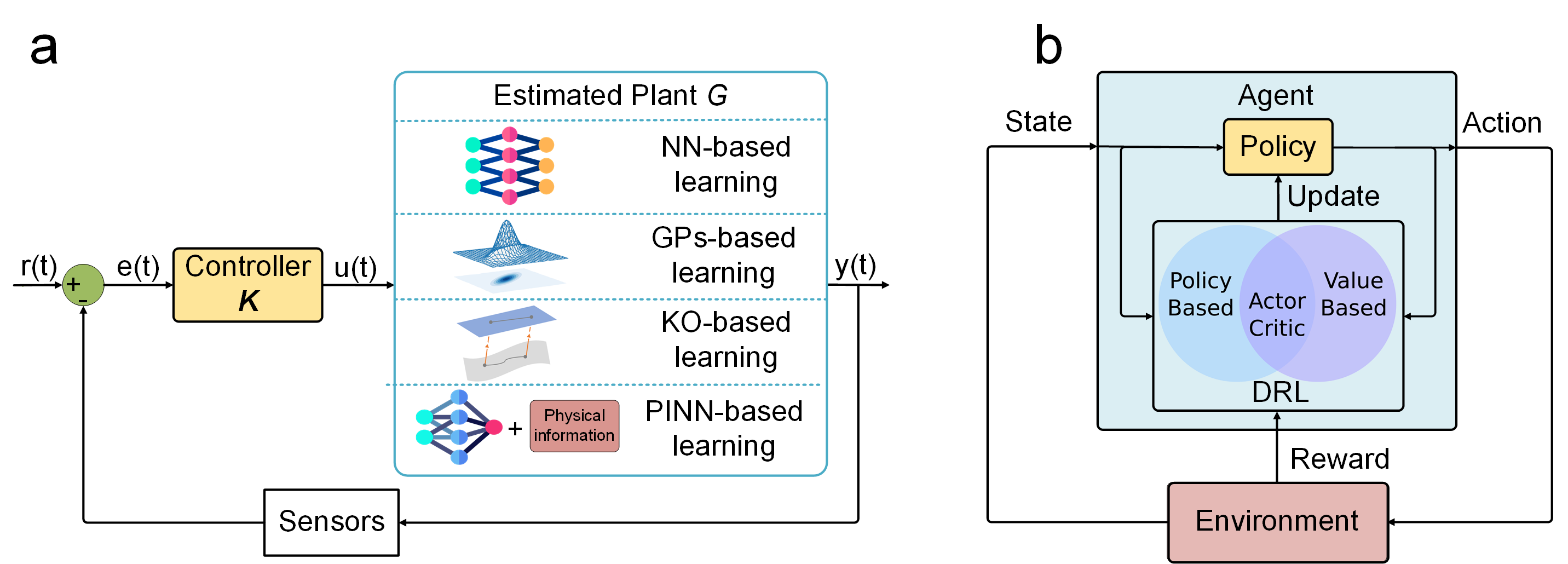}
\caption{Data-driven control schemes for marine robots. (a) Model-based data-driven control, (b) Model-free data-driven control.}
\label{DDMB}
\end{figure}

Model-based data-driven control methods aim to learn approximate system dynamics of marine robots from data, and then use these learned models for controller design. Based on the approach used for dynamic model approximation, these methods can be classified into four categories:
\emph{1)} Neural networks (NNs)-based control methods,
\emph{2)} Gaussian processes (GPs)-based control methods, 
\emph{3)} Koopman operator (KO)-based control methods, and 
\emph{4)} Physical informed neural network (PINN)-based control methods,
as shown in Figure~\ref{DDMB} (\emph{a}).

\subsubsection{Neural Networks (NNs)-Based Control Methods}
NNs, as a universal function approximator, are well-suited for modeling highly nonlinear system dynamics. In marine robot control, NNs are used to learn either forward dynamics or to approximate unmodeled dynamics, uncertainties, and external disturbances using data from simulations or real-world experiments. These learned models have been successfully integrated into control frameworks such as sliding mode control (SMC)~\cite{jiang2022neural} and adaptive control~\cite{8714020,zhang2022neural}, thereby improving the robustness and adaptability of marine robot control systems.
Among various NN architectures, radial basis function (RBF) networks are particularly popular due to their simplicity, localized learning properties, and fast convergence. However, their practical deployment presents several challenges:
\emph{1)} RBF networks are affected by the “curse of dimensionality”—the number of required neurons grows rapidly with input dimensionality, resulting in greater computational demands that complicate real-time execution on embedded platforms.
\emph{2)} The performance of RBF networks is highly sensitive to the placement of centers and the selection of width parameters; suboptimal choices can lead to underfitting, overfitting, or poor control performance.
\emph{3)} Most implementations use a fixed network structure, which may be insufficiently flexible to adapt to dynamic marine environments.

\subsubsection{Gaussian Processes (GPs)-Based Control Methods}
The system dynamics of marine robots can generally be decomposed into a known nominal component and an unknown residual uncertainty term: 
\begin{equation}
\dot{x}=f_{\text {nominal}}(x, u)+\underbrace{f_{\text {residual}}(x, u)}_{\text {learned by GPs}} \label{GP}
\end{equation}
where ${f_{\text {residual}}(x, u)}$ represents unmodeled dynamics, parameter uncertainties, and external disturbances.
For the estimation of ${f_{\text {residual}}(x, u)}$, GPs provide two key advantages over traditional approaches such as computational fluid dynamics or observer-based estimation: \emph{1)} GPs provide a probabilistic framework that captures modeling uncertainty and yields confidence intervals for predictions, and \emph{2)} GPs enable direct, data-driven modeling of unknown dynamics without requiring extensive prior knowledge.
In practical applications, GPs are used to learn the residual uncertainty term and are combined with nominal models based on physical laws or simplified dynamics to construct an enhanced system model. This composite model can then be integrated into advanced control frameworks such as MPC~\cite{li2024gaussian} and SMC~\cite{lima2020sliding}, where the GP’s mean prediction is used for state estimation and its uncertainty quantification informs robust, risk-aware control decisions.
Despite their strengths, GPs face limitations in real-time applications due to high computational complexity and static hyperparameter settings. To address this issue, an incremental sparse GP (ISGP)-based MPC framework was proposed in~\cite{dang2025incremental} for AUV trajectory tracking, using sparsification to discard low-information data and reduce computational load while maintaining model fidelity.
Another limitation of GPs is their reliance on offline hyperparameter tuning, which limits adaptability to dynamic conditions. To overcome this issue, ~\cite{amer2025empowering} introduced the Dynamic Forgetting GP (DF-GP) framework, which enables online adaptation to environmental disturbances without hyperparameter fine-tuning.
An emerging research direction within this framework involves integrating GPs with reinforcement learning (RL)~\cite{cui2022filtered}, where GPs can provide prior knowledge and model rewards to enhance sample efficiency.

\subsubsection{Koopman Operator (KO)-Based Control Methods}
Marine robots often exhibit highly nonlinear dynamics due to fluid–structure interaction (FSI), parameter uncertainties, environmental disturbances, and actuator saturation. The KO provides a powerful framework for modeling such complex systems by lifting the state $\mathbf{x}_k$ into a higher-dimensional space of observables $f\left(\mathbf{x}_{k+1}\right)$, where the system dynamics can be approximated linearly:
\begin{equation}
f\left(\mathbf{x}_{k+1}\right) = \mathcal{K} f\left(\mathbf{x}_k\right)
\end{equation}
Here, $f(\cdot)$ denotes observable functions and 
$\mathcal{K}$ is the (potentially infinite-dimensional) KO.

Due to its ability to represent nonlinear systems through a globally linear formulation, KO-based modeling has gained traction in marine robot control.  It has been successfully applied to both rigid-body platforms such as AUVs~\cite{rahmani2024enhanced} and USVs~\cite{li2024c3d}, as well as to soft-bodied marine robots~\cite{mamakoukas2021derivative}.
To facilitate effective marine robot control, the dynamic models learned via the KO are often integrated with established control methods such as SMC~\cite{rahmani2024enhanced}, MPC~\cite{cong2025koopman}, and linear–quadratic
regulator (LQR)~\cite{li2024c3d,mamakoukas2021derivative}. 
For real-world deployment, adaptability to time-varying dynamics is essential. 
To this end,~\cite{li2024c3d} proposed a cascade control framework for USV station-keeping, using KO for dynamics linearization and a change-point detection module to trigger retraining when system drift was detected.
For real-time control of soft robotic fish, \cite{mamakoukas2021derivative} introduced a KO-based control method that formalized basis function selection and provided theoretical guarantees on approximation quality. 
 A particularly promising direction emerging from this work is the derivative-based synthesis of KOs, which enables effective control of marine robots even in the presence of unknown or partially known dynamics.

\subsubsection{Physical Informed Neural Network (PINN)-Based Control Methods}
PINNs incorporate physical laws, which are typically formulated as partial differential equations (PDEs), into the structure or loss function of NNs. This integration enables the learned models to inherently satisfy the governing equations and physical constraints, leading to improved generalization and interpretability in system modeling~\cite{raissi2019physics}.
Compared to conventional NNs-based methods, PINNs offer two main advantages: \emph{1)} they embed domain-specific physical knowledge into the learning process, ensuring physically consistent predictions and better generalization, and \emph{2)} they can estimate solutions in regions with sparse or incomplete data, making them particularly suitable for marine robot control.
For practical implementation,~\cite{liu2024research} integrated PINN-based dynamic modeling with MPC to achieve trajectory tracking for AUVs. By embedding physics-informed constraints during training, the model achieved improved long-horizon prediction and real-world performance. In~\cite{majumder2024safe}, PINNs were used for safe AUV navigation, incorporating obstacle avoidance through control barrier functions (CBFs). The resulting quadratic programming (QP) formulation was transformed into a Hamilton–Jacobi–Bellman (HJB) equation and solved using a PINN.
Despite their promise, PINN-based control methods face several challenges, including high computational cost, sensitivity to network architecture and hyperparameter settings, and limited scalability to high-dimensional systems. Future research should focus on adaptive physics-based regularization and strategies for real-time deployment.

\subsection{Model-Free Data-Driven Control Methods}
Model-free data-driven control methods are gaining increasing traction in the control of marine robots, providing end-to-end solutions that rely on simulation or experimental data and reward functions instead of relying on explicit system models. 
Existing model-free data-driven control methods can be broadly categorized into: \emph{1)} deep reinforcement learning (DRL), \emph{2)} imitation learning (IL), and \emph{3)} deep imitation reinforcement learning (DIRL).

\subsubsection{Deep Reinforcement Learning (DRL)}
DRL has emerged as a powerful framework for controlling marine robots in complex tasks. By integrating deep neural networks (DNNs) with RL, DRL enables marine robots to learn control policies directly from high-dimensional sensory inputs (e.g., camera, sonar) through trial-and-error interaction with the environment. Typically, DRL problems are modeled as a Markov Decision Process (MDP), defined by a tuple \( (\mathcal{S}, \mathcal{A}, \mathcal{P}, \mathcal{R}, \gamma) \).
\begin{marginnote}[]
\entry{MDP}{Markov Decision Process, defined by a tuple \( (\mathcal{S}, \mathcal{A}, \mathcal{P}, \mathcal{R}, \gamma) \). In which \( \mathcal{S} \) is the set of observable states, \( \mathcal{A} \) the set of possible actions, \( \mathcal{P}(s'|s,a) \) the state transition probability function, \( \mathcal{R}(s,a) \) the reward function, and \( \gamma \in [0,1) \) the discount factor.}
\end{marginnote}
The goal of DRL methods is to learn a policy \( \pi: \mathcal{S} \rightarrow \mathcal{A} \) that maximizes the expected discounted cumulative reward, defined as: 
\begin{equation}
    J(\pi) = \mathbb{E}_{\pi} \left[ \sum_{t=0}^{\infty} \gamma^t r_t (s_t, a_t) \right]
\end{equation}
where \( r_t (s_t, a_t) \) is the reward received at time step \( t \), and the expectation is taken over the trajectories generated by following policy \( \pi \). 

In DRL, DNNs are used to approximate key components of the RL process, such as the value function, policy, or both. Accordingly, DRL methods for the control of marine robots are generally classified into: \emph{1)} value-based methods, \emph{2)} policy-based methods, and \emph{3)} actor-critic methods, as shown in Figure~\ref{DDMB} (\emph{b}).
    
\paragraph{Value-Based Methods} Value-based methods aim to learn a value function that estimates the expected cumulative reward for taking action \( a \) in state \( s \). The optimal policy is derived implicitly by selecting actions that maximize the value \( V^{\pi}(s) \) or action-value function \( Q^{\pi}(s, a) \). These methods are particularly effective in discrete action spaces and are widely applied in tasks such as grid-based navigation and collision avoidance.
In marine robot control, Deep Q-Learning (DQN)~\cite{algarin2025intelligent} and its extensions, including Double DQN (DDQN)~\cite{higo2023development}, Dueling DQN~\cite{wu2020autonomous}, and Dueling Double DQN (D3QN)~\cite{chen2024deep}, have been successfully applied to tasks such as trajectory tracking, autonomous navigation, intelligent collision avoidance, and motion stabilization in marine environments.
Despite these advances, several challenges remain:~\emph{1)} many studies rely on fixed or overly simplified training environments, limiting the real-world applicability of the learned policies;~\emph{2)} evaluations often omit the use of multiple random seeds, which is critical for assessing robustness and reproducibility; and~\emph{3)} practical constraints, such as robot maneuverability and compliance with maritime regulations (COLREGs), are frequently underrepresented in the models.

\paragraph{Policy-Based Methods} Policy-based methods optimize policy parameters 
\(\theta\) directly via gradient-based techniques to maximize expected cumulative reward~\cite{wang2022deep}, making them well-suited for continuous or high-dimensional action spaces.
Proximal Policy Optimization (PPO) algorithm is among the most widely adopted policy-based algorithms for control of marine robots, which has been successfully applied to tasks such as dynamic positioning of USVs~\cite{overeng2021dynamic} and trajectory tracking of AUVs~\cite{hasankhani2023integrated}. However, standard PPO may suffer from sample inefficiency, poor performance under sparse rewards, and sensitivity to hyperparameters—issues that are exacerbated in dynamic marine environments.
To overcome these limitations, recent PPO variants for control of marine robots converge on three strategies: \emph{1)} reward shaping with domain cues to mitigate sparse feedback~\cite{wu2023deep,huang2024learning}, \emph{2)} architecture or training refinements to accelerate convergence and improve stability~\cite{hao2024target}, and \emph{3)} adaptive mechanisms, such as rollback clipping, demonstration replay, or dynamic reward shaping, to boost generalization in nonlinear, current-disturbed settings~\cite{zhang2023auv,chu2025adaptive}.
While each study emphasizes a different aspect, collectively they demonstrate that task-specific PPO adaptations can enable robust, real-time control for navigation, docking, manipulation, and dynamic positioning of marine robots in complex environments.

\paragraph{Actor-Critic Methods} Actor–critic methods combine value-based and policy-based approaches within a unified framework, offering reduced variance in policy gradient estimates and improved convergence. The ability to handle high-dimensional inputs, unmodeled dynamics, parameter uncertainties, and actuator saturation makes actor–critic methods well-suited for the control of marine robots.
Actor–critic methods have been applied to trajectory tracking of USVs~\cite{wang2021data} and AUVs~\cite{deng2021event}, as well as visual servoing control of UVMSs~\cite{wang2024reinforcement}. However, the actor–critic framework often suffers from sample inefficiency and sensitivity to hyperparameter settings. To address these limitations, several advanced variants, such as Soft Actor-Critic (SAC), Deep Deterministic Policy Gradient (DDPG), and Twin Delayed DDPG (TD3), have been proposed to enhance sample efficiency, improve exploration stability, and increase robustness in marine robot control.
\begin{marginnote}[]
\entry{Actor-Critic}{Actor-Critic methods contain two components: an \textbf{actor} that determines which actions to take according to a policy function, and a \textbf{critic} that evaluates those actions according to a value function. 
Specifically, the actor learns the policy \( \pi_\theta(a|s) \), and the critic estimates either the value function \( V^{\pi}(s) \), the action-value function \( Q^{\pi}(s, a) \), or a combination of both.}
\end{marginnote}

$\blacksquare$
\textbf{SAC} is an off-policy actor–critic algorithm that augments the reward objective with an entropy term to encourage stochastic policy learning. This formulation promotes a balanced exploration–exploitation trade-off, making SAC particularly well-suited for the control of marine robots.
SAC has demonstrated strong control performance across a variety of applications, including path following and pose tracking for BURs~\cite{wang2024learning,ma2023position}, AUV trajectory tracking~\cite{wang2025expert}, and USV dynamic positioning~\cite{yuan2023deep}. 
However, practical deployment of SAC remains hindered by challenges such as high computational demand, exploration efficiency, and limited adaptability to dynamic marine environments.
To address these issues, recent studies have introduced several improvements: non-policy sampling strategies to enhance exploration~\cite{dong2025end}, expert-guided learning to reduce data requirements in early training~\cite{wang2025expert}, prioritized experience replay to increase sample efficiency~\cite{yuan2023deep}, and the Sample-Observed SAC (SOSAC) algorithm for improved learning quality and reward accumulation~\cite{ma2023sample}.

$\blacksquare$ 
\textbf{DDPG} avoids action discretization by directly learning a deterministic policy over continuous action spaces. It employs target networks and experience replay in DQN to enhance training stability and sample efficiency, making it particularly effective for high-dimensional control tasks in marine robots.
However, DDPG still faces several challenges in marine robot control, including sample inefficiency, training instability, and limited robustness to dynamic and uncertain environments. To overcome these challenges, enhanced variants have pursued several shared strategies:
\emph{1)} Improving sample efficiency and stability via optimized replay buffers and critic structures~\cite{sun2020auv}, or supervised architectures with prioritized replay~\cite{9351698};
\emph{2)} Reward shaping and constraint design to address underactuation and control limitations~\cite{zhang2022auv};
\emph{3)} Incorporating temporal and contextual awareness through LSTM networks~\cite{masmitja2023dynamic} or Lyapunov-based safety guarantees~\cite{du2022safe};
\emph{4)} Expanding sensory modalities, as seen in vision-based navigation~\cite{zhu2022autonomous};
Hybridization with other DRL methods, notably combining DDPG with SAC for mapless navigation in dynamic settings~\cite{grando2021deep}.
These advancements underscore a common recognition: while DDPG provides a strong baseline for continuous control, its real-world effectiveness in marine robots depends on addressing its inherent limitations through targeted algorithmic innovations.

$\blacksquare$
\textbf{TD3} enhances DDPG by addressing common sources of instability in continuous control. It incorporates clipped double Q-learning to reduce overestimation bias, delayed policy updates for improved training stability, and target policy smoothing to lower variance, making it well-suited for marine environments with unpredictable dynamics.
TD3 has been successfully applied to motion planning and obstacle avoidance for AUVs~\cite{hadi2022deep}. However, two key challenges remain in its practical implementation. The first is high-frequency action oscillation, which can lead to actuator wear. To mitigate the action
oscillation, policy regularization and adaptive reward shaping have been employed to promote smoother control actions~\cite{fan2024path}. The second challenge involves inefficient experience replay. To address this issue, hybrid-priority replay strategies that combine temporal-difference (TD) error and reward magnitude have been proposed~\cite{zhou2025adaptive} to enhance sample efficiency.
Comparative evaluations of DDPG, TD3, SAC, and PPO under varying sea conditions have provided valuable guidance for selecting appropriate DRL algorithms for marine control tasks~\cite{sivaraj2023performance}, with TD3 demonstrating a strong balance of robustness and precision when properly tuned.

\subsubsection{Imitation Learning (IL)}
Reward design remains a fundamental challenge in DRL, as poorly defined or sparse rewards can impede learning and result in suboptimal policies. Moreover, hand-crafted, task-specific reward functions often lack generalizability, particularly in dynamic or previously unseen marine environments. IL addresses these limitations by enabling marine robots to acquire control policies directly from expert demonstrations, thereby reducing or eliminating the need for explicit reward engineering.
Common IL approaches designed for the control of marine robots include:
\emph{1)}~Behavior Cloning (BC),
\emph{2)}~Inverse Reinforcement Learning (IRL), and
\emph{3)}~Generative Adversarial Imitation Learning (GAIL).

\paragraph{Behavior Cloning (BC)} BC is a fundamental IL approach in which an agent learns a direct mapping from observations to actions using supervised learning. In marine robot control, BC has been applied to goal-conditioned visual navigation for AUVs~\cite{manderson_vision-based_2020}, where expert-labeled image frames were used to guide pitch and yaw commands for tasks like obstacle avoidance and reef-top following.
Recent advancements include UIVNAV~\cite{lin2024uivnav}, an end-to-end control system for obstacle-avoiding underwater navigation over objects of interest. It bypasses explicit localization by leveraging domain-invariant visual representations and human-annotated control inputs. To address limitations in policy refinement and performance saturation, AquaBot~\cite{liu2024self} combined BC with self-learning, enabling autonomous improvement beyond teleoperated demonstrations in underwater robotic manipulation tasks. Within this topic, future research directions include dataset aggregation, domain randomization, and hybrid IL–RL frameworks aimed at enhancing robustness, generalization, and adaptability in complex and dynamic marine environments.

\paragraph{Inverse Reinforcement Learning (IRL)} IRL aims to infer the underlying reward function from expert demonstrations, enabling the use of standard RL algorithms to learn policies that generalize beyond the observed behavior. In marine robot control, IRL is particularly valuable in sparse-reward settings where manual reward design is challenging. For example,~\cite{zhang2022leveraging} proposed DDPG-DIR, an IRL-augmented DDPG framework for pose regulation of a robotic fish with multiple objectives. To support demonstration collection from non-expert users, this method introduced a DMP-TDG to generate diverse trajectories from a single example. The effectiveness of DDPG-DIR was validated through real-world experiments.
Despite its promise, IRL faces challenges such as high computational cost and reward ambiguity, where multiple reward functions may explain the same behavior. These limitations are further amplified in dynamic and partially observable marine environments. Future work should focus on enhancing scalability and robustness of IRL through integration with generative modeling, meta-learning, or online adaptation mechanisms.

\paragraph{Generative Adversarial Imitation Learning (GAIL)} GAIL combines Generative Adversarial Networks (GANs)~\cite{goodfellow2020generative} with IL to implicitly learn control policies without explicitly recovering a reward function. It simultaneously trains a discriminator to differentiate between expert behavior and agent behavior, and a generator (policy) to produce actions that mimic expert demonstrations. This adversarial framework enables effective policy learning while avoiding the challenges associated with manual reward engineering.
In~\cite{chaysri2023unmanned}, GAIL was applied to USV navigation under environmental disturbances, demonstrating improved control performance and generalization across diverse marine conditions. ~\cite{jiang2022generative} introduced Generative Adversarial Interactive Imitation Learning (GA2IL) for AUV path-following. GA2IL extends GAIL by incorporating interactive human feedback, which was interpreted as evaluative signals based on expert knowledge. This hybrid framework enabled the marine robot to refine its behavior during training, thereby enhancing robustness in complex environments.
These studies underscore the potential of GAIL-based approaches to learn high-performance, generalizable control policies in dynamic marine settings.

\subsubsection{Deep Imitation Reinforcement Learning (DIRL)}
DIRL is a hybrid framework that combines the strengths of IL and DRL, in which IL provides rapid policy initialization from expert demonstrations, while DRL enables policy refinement through trial-and-error interaction, thereby enhancing adaptability and generalization.
A basic DIRL framework was investigated in~\cite{chu2020motion} for AUV motion control by integrating IL with DDPG and TD3. Among the approaches tested, IL-TD3 outperformed others, achieving double the convergence speed and improved training stability compared to standalone DRL methods.
To improve real-world applicability,~\cite{yan2022real} proposed a DIRL framework for underwater exploration with a robotic fish. The framework comprised three components: initial policy learning via IL from expert data, online fine-tuning through DDPG, and a consolidated IL module. Inspired by game mechanics, the system stored the highest-reward trajectory from each episode in an imitation buffer, which was periodically used for offline IL.

\subsection{Integration of Model-Based and Data-Driven Control Methods}
\label{sec:Integration }
Model-based control methods (e.g., SMC, MPC) provide interpretability, formal stability guarantees, and explicit handling of physical constraints. However, their effectiveness depends heavily on the accuracy of system models, which can be difficult to obtain for marine robots operated in highly dynamic environments.
Conversely, data-driven approaches, especially RL, excel at learning unknown dynamics and coping with unstructured environments, yet lack theoretical guarantees and interpretability.
To address these limitations, hybrid controllers that combine model-based and data-driven methods have emerged. These hybrid controllers aim to integrate the rigor and reliability of model-based frameworks with the adaptability of data-driven techniques.

\subsubsection{Integration of MPC and RL}
Integrating MPC with RL leverages the constraint-handling and optimization strengths of MPC alongside the adaptability of RL. This hybrid paradigm has shown superior performance in marine robotic tasks such as trajectory tracking, path following, station-keeping, and target reaching, especially under uncertain and partially observable conditions.
In RL-enhanced MPC frameworks, RL is typically employed to learn system dynamics or cost functions. For instance, Q-learning has been combined with nonlinear MPC to facilitate real-time adaptation to disturbances~\cite{martinsen2022reinforcement}, while~\cite{zhang2024deep} employs data-driven models obtained during exploration to support predictive control planning. To address partial observability, Bayesian filtering and GPs-based uncertainty modeling have been introduced to enhance control performance~\cite{cui2022filtered}.
More advanced architectures, such as actor–model–critic (AMC) frameworks, incorporated learned dynamics to enhance spatiotemporal reasoning~\cite{ma2023neural}. In~\cite{cui2021autonomous}, probabilistic MPC informed by RL demonstrates high sample efficiency and adaptability. Additionally, parameterized MPC schemes guided by DDPG enable policy refinement without the need for manual tuning~\cite{pan2022learning}.
An emerging trend reverses this integration, using MPC to guide, constrain, or initialize RL policies.

\subsubsection{Integration of SMC and RL}
The integration of SMC with RL combines the robustness and stability guarantees of SMC with the adaptability and model-free learning capabilities of RL. In such hybrid frameworks, RL is typically employed to tune switching gains, estimate unmodeled dynamics, or learn sliding surfaces, thereby improving control performance in dynamic marine environments.
Recent studies demonstrate the promise of this integrated framework. DDPG has been used to automate SMC parameter tuning for AUV motion control across varying speeds~\cite{8734349}, while in~\cite{zhu2025sliding}, TD3 enabled real-time parameter optimization in adaptive SMC for USV trajectory tracking under disturbances. Actor–critic methods have also been applied to heading control of AUVS, learning optimal SMC parameters under system uncertainty~\cite{wang2022sliding}.
Promising research directions include the integration of Lyapunov-based safe RL, the reduction of chattering through smooth policy approximations, and the real-time deployment in marine robotic systems.

\section{COOPERATIVE CONTROL OF MULTIPLE MARINE ROBOTS} 
Motivated by the cooperative behaviors observed in biological collectives, extensive research has focused on the cooperative control of marine robots. By enabling coordinated behavior among multiple marine robots, cooperative control allows the overall system to achieve synergistic effects, where the collective performance exceeds the sum of individual capabilities.

\subsection{Cooperative Scenarios of Multiple Marine Robots}
\begin{figure}[htb]
\centering
\includegraphics[width=6.2in]{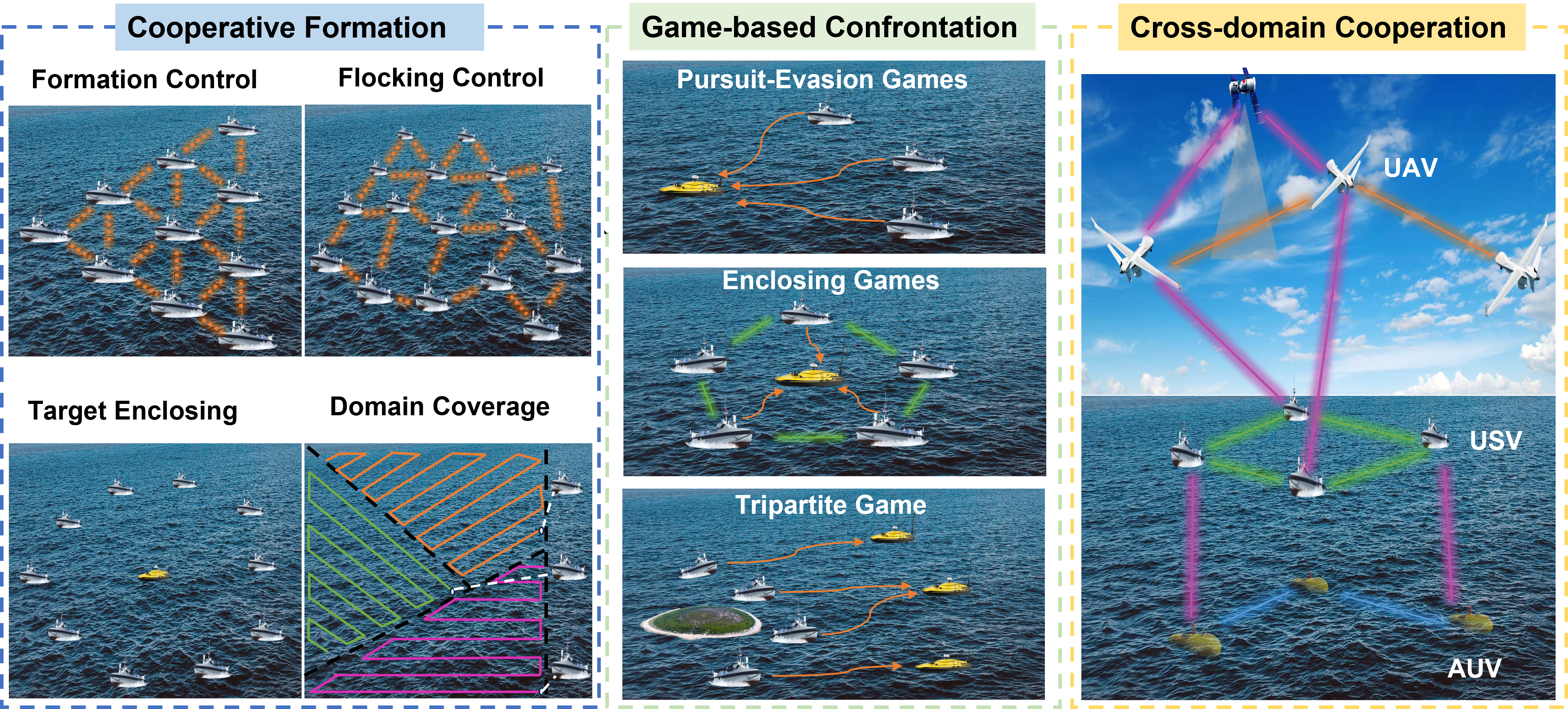}
\caption{Deployment scenarios of cooperative multiple marine robots.}
\label{scenario}
\end{figure}

Marine robots are capable of executing a variety of tasks through cooperative control strategies. Based on task objectives, cooperative control methods can be broadly classified into three categories: \emph{1)} cooperative formation,  \emph{2)} game-based competition, and  \emph{3)} cross-domain coordination, as illustrated in \textbf{Figure~\ref{scenario}}.

\subsubsection{Cooperative formation of multiple marine robots}
Cooperative formation control aims to enable multiple marine robots to achieve efficient and coordinated motion with formation coordination by leveraging inter-agent communication, distributed decision-making, and cooperative control strategies. In this context, formation coordination is categorized into: \emph{1)} formation control, \emph{2)} flocking control, \emph{3)} target encirclement, and \emph{4)} area coverage.

\paragraph{Formation Control}
Formation control focuses on maintaining predefined geometric configurations, such as line-abreast, column, or triangular patterns, among multiple marine robots during cooperative operations~\cite{Peng-TC-2020}. In practical tasks, the objective is to preserve formation cohesion while adapting to environmental disturbances (e.g., obstacle avoidance) and task-specific requirements (e.g., formation switching).
Based on motion guidance strategies, formation control methods are generally categorized into three types: trajectory tracking~\cite{Jiang-OE-2024-Adaptive}, path following~\cite{Guo-AOR-2024-adaptive}, and target tracking control~\cite{Ma-JAS-2023}.
Despite notable advancements in algorithm development and field deployment, key challenges remain, particularly ensuring robustness under communication constraints, achieving scalability for large formations, and maintaining adaptability in dynamic marine environments.

\paragraph{Flocking Control}
Flocking control aims to replicate the self-organizing behaviors seen in biological groups such as fish schools and bird flocks, enabling multiple marine robots to make autonomous decisions, maintain coordinated formations, and execute tasks efficiently—enhancing adaptability and performance in complex environments.
Flocking control methods can be broadly categorized into:
\emph{1)} control in static environments, and
\emph{2)} control in dynamic environments.
In static environments, where factors like obstacle locations are fixed and known, flocking control focuses on cooperative motion planning and formation maintenance using global path planning and fixed-formation strategies~\cite{Sahu-TFS-2018}.
In dynamic environments, the focus shifts to real-time adaptation—such as obstacle avoidance and flexible formation adjustments—based on local sensing, adaptive control, and dynamic reconfiguration. These methods handle challenges like moving obstacles, time-varying disturbances, and changing communication topologies~\cite{Peng-TSMCS-2023,Peng-TC-2021}.
Despite promising results in simulation, flocking control still faces challenges in real-world deployment. Bridging this gap remains a critical direction for future research.

\paragraph{Target Enclosing} 
Target-enclosing control coordinates multiple marine robots to form and maintain a stable formation around one or more moving targets, while meeting requirements on inter-robot spacing, angular positioning, and safety margins under environmental disturbances and model uncertainties.
Based on the number of targets involved, target enclosing control can be categorized into:
\emph{1)} Single-target enclosing control \cite{Zhang-OE-2025};
\emph{2)} Multi-target enclosing control \cite{Jiang-TFS-2024}.
Based on the availability of target information, enclosing strategies can be divided into:
\emph{1)} Enclosing control with known target information \cite{Zhang-OE-2025}, where robots have access to real-time target states (e.g., position or velocity). 
\emph{2)} Enclosing control with unknown target information \cite{Yan-ASME-2025}, which requires online estimation of target states and compensation for environmental disturbances such as ocean currents or wave effects. 
The former enclosing strategy is suitable for scenarios with reliable communication and high-fidelity sensor feedback, whereas the latter is more appropriate for scenarios characterized by limited communication or partial observability.

\paragraph{Domain Coverage}
Domain coverage requires coordination among multiple ocean robots to execute planned coverage paths to effectively scan designated areas and reasonably avoid obstacles along the path. The objective of domain coverage control is to achieve uniform, efficient, and adaptive coverage through cooperative marine robot control, which is critical for missions such as search, reconnaissance, and environmental monitoring.
Based on robot mobility and environmental dynamics, domain coverage control strategies are generally classified into:
\emph{1)} Static coverage~\cite{Luo-OE-2024}, emphasizing optimal static deployment, typically using Voronoi partitioning to uniformly allocate robots across a predefined area;
\emph{2)} Dynamic coverage~\cite{Jiao-OE-2025, Liu-TVT-2025}, which leverages robot mobility to adaptively modify coverage patterns in response to changes in the environment or mission objectives.
 For effective dynamic coverage control, robots must perceive their surroundings within constrained sensing ranges while ensuring safe navigation through both static and dynamic obstacles.

\subsubsection{Game-Based Competition of Multiple Marine Robots}
Game-based competitive cooperation among multiple marine robots involves coordinated execution of tasks such as dynamic pursuit, encirclement, and target expulsion. These tasks are governed by collaborative decision-making frameworks grounded in game theory, which facilitate strategic interactions in adversarial or semi-cooperative scenarios.
Based on the operational context, game-theoretic tasks in marine robotics are commonly classified into:
\emph{1)} Pursuit–evasion games,
\emph{2)} Enclosure–capture games, and
\emph{3)} Tripartite adversarial games.

\paragraph{Pursuit-Evasion Games}
Pursuit–evasion games model intelligent adversarial interactions between groups of marine robots—pursuers and evaders—operating in complex underwater or surface environments with multiple obstacles~\cite{Li-TSMCS-2024,Meng-TAI-2024}. Pursuers collaboratively generate interception paths to capture the evader or constrain its maneuvering space as efficiently as possible while avoiding collisions. Meanwhile, the evader reacts strategically to the pursuer's movements to evade capture or extend the duration of pursuit.
This game-theoretic framework is widely applicable in real-world maritime operations, including anti-smuggling patrols, anti-submarine warfare, and enforcement against illegal fishing.

\paragraph{Enclosing Games}
Enclosing games model dynamic adversarial interactions between marine robot teams (encirclers) and targets in complex aquatic environments~\cite{Jiang-TIV-2024, Nantogma-Ele-2023}. Encirclers employ cooperative strategies derived from non-cooperative game theory to design distributed control algorithms that enable rapid convergence, formation of symmetric enclosures, and real-time adjustments to maintain equidistant spacing, while avoiding collisions with obstacles and neighboring agents.
Meanwhile, the target executes evasive maneuvers to escape encirclement. This game-theoretic framework underpins a range of maritime applications, including target interception, surveillance and blockade operations, and the containment of hostile or unauthorized marine robots.

\paragraph{Tripartite Games}
Tripartite games model dynamic adversarial interactions among three parties: attackers (e.g., torpedoes), defenders (e.g., interceptors), and high-value targets~\cite{Kang-TIE-2024,Du-OE-2022}.
Each party operates within a shared environment while pursuing distinct objectives. The target, typically a naval vessel or coastal facility, is usually passive, with its primary aim being to avoid intrusion or damage.
Defenders employ cooperative strategies such as dynamic path planning, real-time obstacle avoidance, and interception control to shield the target while ensuring their own safety. In contrast, attackers seek to breach the defended zone or disable the target, often using evasive maneuvers to reduce interception risk.
This game-theoretic framework supports critical applications, including maritime infrastructure protection, military zone defense, and anti-smuggling enforcement.

\subsubsection{Cross-Domain Cooperation of Multiple Marine Robots}
Cross-domain cooperation refers to the coordinated operation of heterogeneous unmanned systems, such as USVs, AUVs, and and unmanned aerial vehicles (UAVs), operated across different domains under unified mission objectives. Despite variations in platform capabilities and operating environments, these systems achieve integrated performance through complementary functions, real-time information sharing, and dynamic task allocation.
The primary goal is to leverage multi-domain resources to enhance communication, situational awareness, real-time decision-making, and coordinated control in dynamic environments, thereby improving overall mission efficiency and responsiveness.
Cross-domain cooperation is especially valuable for tasks requiring multi-dimensional spatial coverage and rapid adaptability. Based on spatial interaction types, it can be classified into:
\emph{1)} Air–underwater cooperation~\cite{Feng-ELL-2025};
\emph{2)} Surface–underwater cooperation~\cite{Jiang-SSCAE-2024};
\emph{3)} Air–surface–underwater cooperation~\cite{Cao-TIV-2024}.
Typical applications include marine environmental monitoring, military surveillance, and civilian missions such as search and rescue or offshore infrastructure inspection.

\subsection{Cooperative Control of Multiple Marine Robots}
Model-based methods have long dominated the cooperative control of marine robots, offering formal stability guarantees and interpretable architectures grounded in system dynamics and control theory. However, these methods often face limitations in handling unmodeled dynamics, parameter uncertainties, and external disturbances commonly encountered in real-world marine environments. In response, data-driven control schemes have attracted growing attention due to their adaptability. A summary of model-based cooperative control methods for multi-marine-robot systems is provided in \textbf{Supplemental Material 2}.

\begin{marginnote}[]
\entry{\href{https://zenodo.org/records/15731778?token=eyJhbGciOiJIUzUxMiJ9.eyJpZCI6IjE1OWQ0YmU5LTVkYTktNGFjZi1iOTUyLTRlOTI5MWVhYmVkYiIsImRhdGEiOnt9LCJyYW5kb20iOiIzNjA1ZDBjN2U4OWNkMGE3NzM2YzYyOWFiMjEyNTFmNCJ9.XdNXNIjRJV6vmFTd4vgk8-VB2pW-emDIHongXI1G55TCcmlQ9gl5rzEnJ9D2EO2iyQibFYjKoYJ9RXcEvzAbnQ}{Supplemental Material 2}}{A summary of model-based cooperative control methods for multiple marine robots}

\end{marginnote}

\subsubsection{Multi-Agent Reinforcement Learning (MARL)-Based Cooperative Control}
MARL enables marine robots to autonomously learn cooperative or competitive behaviors in dynamic and uncertain environments. For instance, \cite{Liu-JOE-2023} proposed a MARL framework for cooperative search and disturbance-resilient navigation in multi-USV systems. \cite{Wang-OE-2023} integrated RL with backstepping control in a game-theoretic formation strategy to ensure robust tracking performance for multi-AUV systems under environmental disturbances and actuator saturation.
To address constraints imposed by underwater communication, \cite{Cao-JAS-2023} introduced a communication-aware approach that integrates RL with probabilistic modeling to jointly optimize formation stability and link efficiency. An adaptive fixed-time controller was proposed by \cite{Wang-TNSE-2023}, incorporating a quadratic term to bound actor–critic estimation errors and ensure convergence within a fixed time.
Cloud-assisted cooperative control was explored in \cite{Ding-TIV-2023}, where a hierarchical RL framework delegates state estimation to the cloud while retaining a local actor–critic controller for precise tracking, thereby alleviating bandwidth constraints. Furthermore, to defend against cyber threats, \cite{Ding-TFS-2024} developed a secure hierarchical architecture that combines a resilient estimator with a predefined-time fuzzy RL controller, enabling secure surrounding formation within a predefined time horizon.

\subsubsection{Deep Multi-Agent Reinforcement Learning (DMARL)-Based Cooperative Control}
DMARL enables end-to-end policy learning from raw sensory inputs, making it particularly suitable for the control of marine robots in unstructured environments. DMARL approaches are generally classified into cooperative and competitive settings.
In cooperative scenarios, such as obstacle avoidance, agents share common goals and global rewards. These methods have improved decision-making and navigation safety in unstructured marine environments~\cite{Zhang-ITJ-2024,Gan-TASE-2023}. To mitigate COLREG violations and enhance situational awareness,~\cite{Wang-TITS-2025} proposed a collaborative collision avoidance approach that integrates risk assessment with an enhanced deep recurrent Q-network, achieving robust performance in USV fleets.
In competitive scenarios, such as pursuit–evasion and attack–defense, DMARL also demonstrates strong performance. \cite{Li-TSMCS-2024} proposed a distributed strategy for USV swarms with limited perception capabilities, employing multi-agent MAPPO with a bidirectional GRU network to enhance coordination. \cite{Meng-TAI-2024} introduced a cooperative advantage actor–critic framework tailored for distributed pursuit missions. To further support coordination under perception and communication constraints, \cite{Xu-OE-2021} developed a DRL framework based on a Coding-Convolutional Network and centralized actor–critic for multi-AUV systems.

\subsection{Challenges in Cooperative Control of Multiple Marine Robots}  
Coordinated control of marine robots remains challenging due to dynamic environments and unreliable communications. Two major challenges are communication constraints and accurate flow-field estimation. A comprehensive summarization of challenges in cooperative control of multiple marine robots is provided in~\textbf{Supplemental Material 3}. 

\begin{marginnote}[]
\entry{\href{https://zenodo.org/records/15731774?token=eyJhbGciOiJIUzUxMiJ9.eyJpZCI6Ijc4YjY4MDYyLWFkYjktNDZjMi04ZTA5LTYzYzIyZmUyNGQ5ZiIsImRhdGEiOnt9LCJyYW5kb20iOiIxYTcyYWVjOGM5NWFmNzJlZjMxNjczMDI5YWZmYmFlMCJ9.l2q7c7_G1_alxJ6yWYaMakMLiSDkcnPoqipiOKKVKKAP9zAFPLZpqacKVEs-PPtk7DIJ1yZdOY7-OMfFMjCHBA}{Supplemental Material 3}}{A comprehensive summarization of challenges in cooperative control of multiple marine robots}
\end{marginnote}

\subsubsection{Communication Constraints in Marine Robot Cooperation}
Marine communication networks are characterized by low bandwidth, high latency, packet loss, and cyber threats, all of which significantly impair the real-time coordination of multi-marine-robot systems, particularly in large-scale or long-duration missions. Specific communication constraints and their corresponding solutions are summarized as follows:
\emph{1)} Computational burden.  
Distributed control architectures can reduce onboard computational demands and enhance system robustness. Practical applications include guidance and containment for USV path planning~\cite{Peng-TCST-2018}, disturbance-rejection dynamic positioning~\cite{Peng-OE-2022}, multi-layer frameworks for handling model uncertainties~\cite{Lyu-TCYB-2024}, and containment control under time-varying constraints~\cite{Sun-Ne-2022}.
\emph{2)} Limited bandwidth. 
To mitigate communication load while maintaining control performance, event-triggered~\cite{Fan-OE-2025}, self-triggered~\cite{Liu-TIE-2024}, and quantized control strategies~\cite{Wang-TIE-2023} offer effective solutions under bandwidth constraints.
\emph{3)} Communication delay.  
Time-delay estimators~\cite{Suryendu-TCSII-2020} and model-free controllers~\cite{Yan-TCST-2019} have been proposed to ensure the stability and tracking performance of multi-marine-robot systems in the presence of communication delays.
\emph{4)} Network attacks. 
To counter cyber threats such as deception and DoS attacks, several robust methods have been developed, including quantized finite-time control~\cite{Jiang-OE-2024}, fixed-time cooperative strategies~\cite{Gao-TIV-2022}, and adaptive NN controllers capable of handling time-varying intrusion patterns~\cite{Gu-ISA-2020}.
\emph{5)} Packet loss.  
To mitigate the effects of data loss and irregular sampling, non-fragile control protocols~\cite{Jiang-JSYST-2022} and interleaved event-triggered mechanisms~\cite{Zhou-TASE-2025} have been developed to prioritize critical transmissions and improve network reliability.

\subsubsection{Flow Fluid Estimation with Multiple Marine Robots}
The motion of marine robots is strongly influenced by ambient flow, making accurate flow field estimation critical for effective cooperative control of marine robots. Traditional methods for observing flow fields, such as buoy networks or satellite remote sensing, are often limited by high costs and low temporal or spatial resolution. A promising alternative leverages the position and velocity measurements of the robots themselves to estimate local flow conditions, offering a scalable and cost-effective solution. Due to inherent uncertainties in flow modeling, discrepancies between predicted and actual trajectories, known as motion-integration errors, implicitly encode local flow information. Motion tomography (MT)~\cite{wu2013glider,chang2016glider,chang2017motion} exploits these errors to reconstruct high-resolution flow fields. Extensions of MT include spatial–temporal flow decomposition~\cite{chang2016distributed} and distributed estimation via the nonlinear Kaczmarz method~\cite{chang2019distributed}, enabling decentralized flow sensing by multiple robots. These approaches have been validated on platforms such as UGs and UBRs~\cite{zuo2023bio}.

\section{OPEN-SOURCE SIMULATION PLATFORMS AND MARINE ROBOTS}
\subsection{Open-Source Simulation Platforms}
\begin{figure}[htb]
\centering
\includegraphics[width=6.2in]{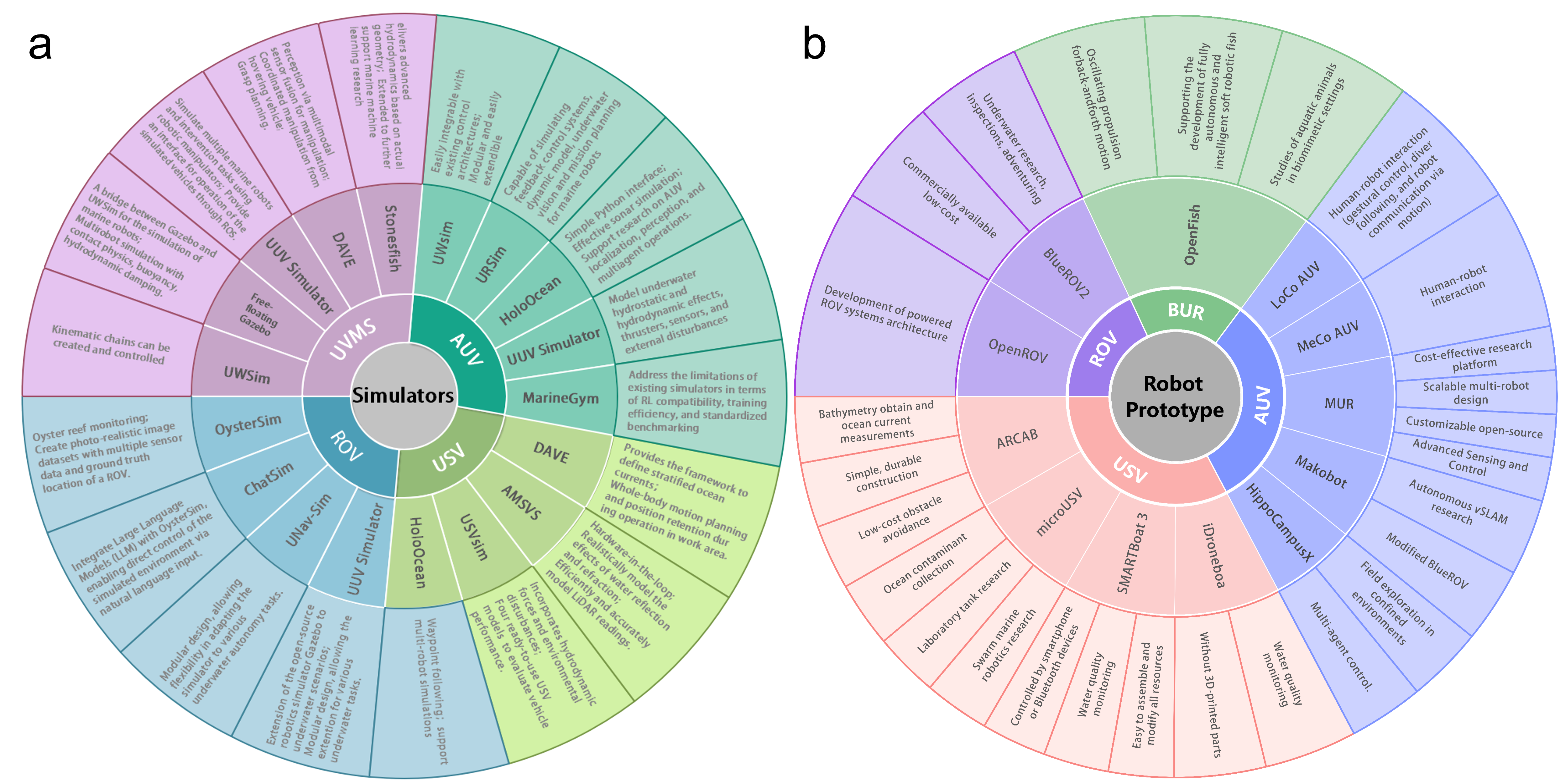}
\caption{Summary of open-source simulation platforms and marine robots.}
\label{opensource}
\end{figure}

Validating control methods for marine robots presents significant challenges due to the high cost of robot prototypes, time-intensive deployments, and associated safety risks. To address these issues, 3D simulation platforms (\textbf{See Figure~\ref{opensource} a}) provide a practical alternative, offering a safe, flexible, and controlled environment for early-stage testing and validation of control strategies.

\subsubsection{Open-Source Simulation Platforms for a Single Marine Robot} 
Simulation platforms for single marine robots are designed to replicate prototype kinematics and hydrodynamic dynamics, while incorporating embedded sensing and motion control capabilities, thereby offering reliable and cost-effective testbeds for validating control strategies (\textbf{See Table 1}).
\begin{marginnote}[]
\entry{\href{https://zenodo.org/records/15731765?token=eyJhbGciOiJIUzUxMiJ9.eyJpZCI6IjcxZDQwYzBhLTdmYzEtNGI0ZC1iNzA3LWFkYmViMTg1YjVkNyIsImRhdGEiOnt9LCJyYW5kb20iOiJkYTc3YmQ4MWJiZGU0ODllZjRlMWM1ZjZjMzY2MzRlNyJ9.0uvlXq4l5U9EPWhUUg6S9dGl2rgH3LYpJWjSRRgSwIKRvNv1uGHL4Olf0nS1_utZg3hj0GyyvkSNmk59zoq9vA}{Supplemental Material 4}}{Table 1}
\end{marginnote}
To this end, existing simulation platforms can be characterized by several key components, including marine robots and environment models, dynamic models, sensor suites, physics engines, and control libraries.

$\blacksquare$ \textbf{Marine robots and environment models}. In most simulation platforms, marine robots are represented by articulated 3D models (e.g., URDF files) that embed their kinematic chains and mass–inertia properties. To recreate the operational context, the marine environment is rendered with Unity3D, Unreal Engine (UE4/UE5) or photorealistic 3D reconstructions generated from sonar or camera, providing visually accurate scenarios.

$\blacksquare$ \textbf{Dynamic models}. Gazebo-based simulators (e.g., Free-Floating Gazebo and UUV Simulator)  incorporate hydrodynamic effects, including added mass, linear and nonlinear damping. USVsim extends this by simulating environmental forces such as waves and ocean currents. Simulation packages for UVMS (e.g., Free-Floating Gazebo, DAVE) also account for dynamic coupling between the base robot and manipulators.

$\blacksquare$ \textbf{Sensor suits}. Standard sensor configurations across simulators include cameras, IMUs, and depth sensors. Sonar is supported in UWSim, HoloOcean, OysterSim, and ChatSim, while LiDAR is support in AMSVS, DAVE, and MARUs. Additional sensor options range from acoustic transponders (e.g., UUV Simulator) to GPS modules (e.g., HoloOcean and USVsim). DAVE further enables multimodal sensor fusion. Range and proximity sensors, available in USVsim, UNav-Sim, and MarineGym, support obstacle avoidance.

$\blacksquare$ \textbf{Control libraries}. PID is commonly supported across simulation platforms. More advanced control strategies, such as MPC and DRL, are available in platforms like UNav-Sim. Most simulators offer compatibility with ROS or ROS 2, facilitating seamless integration with real-world marine robotic systems.

Recent advancements in marine robot simulators reflect a growing emphasis on high-level autonomy. Modern platforms increasingly integrate capabilities such as natural language interfaces and learning-based motion planning, which facilitate intuitive human–robot interaction and support more autonomous task execution. Notably, MarineGym addresses this shift as the first simulator specifically designed for RL. It incorporates GPU-accelerated large-scale parallel simulations, a dedicated RL training pipeline, built-in benchmarking environments, and a domain randomization toolkit to support Sim2Real transfer. Collectively, these developments highlight the expanding role of simulation platforms, not only as testbeds for evaluating control algorithms but also as comprehensive environments for the development and validation of embodied artificial intelligence in marine robotic systems.

\subsubsection{Open-Source Simulation Platforms for Multiple Marine Robots}
Simulation platforms for multiple marine robots are designed to model cooperative behaviors within shared environments, with particular emphasis on inter-robot communication. In addition to the core components found in single-robot simulators, these platforms must simulate realistic communication mechanisms to effectively evaluate the robustness, efficiency, and scalability of coordination strategies.
A prominent example is HoloOcean, developed on UE4, which supports multi-agent underwater scenarios. It offers an extensive sensor suite, including cameras, Doppler Velocity Log (DVL), pressure sensors, GPS, IMU, magnetometers, and sonar, and implements both optical and acoustic communication models.
Despite increasing interest in cooperative control of marine robots, open-source simulators tailored to multi-marine-robot systems remain scarce. This gap highlights the pressing need for scalable, high-fidelity simulation platforms to advance research in multi-marine-robot systems and accelerate real-world deployment of cooperative control methods.

\subsection{Open Source Marine Robots} 
Open-source marine robotic systems (\textbf{See Figure~\ref{opensource} b}) are instrumental in promoting global collaboration by enabling shared development, cross-institutional validation, and community-driven innovation. Publicly available hardware designs and software frameworks of marine robots significantly lower entry barriers for students, researchers, and smaller institutions. The existing open-source marine robots are summarized in \textbf{Table 2}, highlighting their hardware platforms and supported software frameworks.

\begin{marginnote}[]
\entry{\href{https://zenodo.org/records/15731750?token=eyJhbGciOiJIUzUxMiJ9.eyJpZCI6IjYzYzFhZDVhLWFmMDgtNDI1ZS04ODZjLWZjYTEyMzUzNTczYiIsImRhdGEiOnt9LCJyYW5kb20iOiI5N2I4ZDAxNjEyZmJiMmJlMDgwNzMyZGUyNWVjYWU4MSJ9.WPUWL-sSUerZ2cWLoNEID7sUzuBbNoZRFXe_usVzAlUq48tYwYFGNfDIFexQE5sRISz3eNMmZGkZ2p8Kb6dhtA}{Supplemental Material 5}}{Table 2}
\end{marginnote}

\subsubsection{Open Source ROVs}
BlueROV2 is a widely adopted, low-cost, modular open-source ROV equipped with diverse sensors and payloads, including cameras, DVL, sonar, depth sensors, and manipulators. Built on the Pixhawk autopilot and an open-source software stack, BlueROV2 supports customization of both hardware and control algorithms, making it a benchmark platform for underwater robotics research and development. Several simulation environments, such as a Simulink-based setup and UNav-Sim, have been developed around it for algorithm development and validation.
In contrast, OpenROV is a smaller, lightweight open-source alternative with a simpler setup: basic thrusters, a camera, and an arm. While more compact and cost-effective, OpenROV provides limited instrumentation and payload capacity compared to BlueROV2, making it less suited for complex tasks such as precision manipulation or high-resolution mapping in underwater environments.

\subsubsection{Open Source AUVs}
Several open-source AUV platforms have been developed to support accessible and customizable underwater robotics research. LoCO AUV is a compact, low-cost system equipped with tri-thrusters, vision-based algorithm support, and a Gazebo-based simulator, making it well-suited for educational applications. MeCO AUV, featuring a modular aluminum frame, is designed for underwater human–robot interaction and supports real-time deep learning on embedded hardware. MUR is a highly modular, ROS-based AUV, enabling multi-robot scalability,  flexible communication protocols (including acoustic, Wi-Fi, and Ethernet), and advanced perception capabilities using multiple onboard cameras. Makobot HAUV extends the BlueROV2 into a hovering AUV configuration optimized for visual SLAM (vSLAM) research. HippoCampusX is an agile micro-AUV designed for confined and cluttered environments, incorporating a quad-rotor propulsion design, PX4 autopilot, and onboard processing via a Raspberry Pi.

\subsubsection{Open Source USVs}
ARCAB is a portable catamaran platform designed for safe and efficient bathymetric and current data collection. It supports both autonomous and manual operation modes and includes basic obstacle avoidance capabilities. iDroneboat is tailored for real-time, IoT-enabled water quality monitoring, integrating environmental sensors, GPS, LTE communication, and a waterproof enclosure to ensure reliability in field deployments.
SMARTBoat 3 is a compact and ultra-low-cost USV that leverages off-the-shelf sensors and a Bluetooth-enabled Android application for monitoring near-surface water conditions in calm environments. microUSV is a laboratory-scale USV optimized for swarm robotics research. Its single-hull, miniature design supports low-cost multi-agent experiments, making it ideal for studying coordination strategies in constrained environments.

\subsubsection{Open Source UBRs}
OpenFish is an open-source soft robotic fish designed to emulate thunniform swimming through a combination of active and passive tail segments. It utilizes continuous rotation of a DC motor to actuate tensioned cables on either side of the active tail, enabling high-frequency oscillations and generating a more sinusoidal tail trajectory, which results in a maximum swimming speed of 0.85 m/s. This robotic platform provides a testbed for validating models of oscillatory propulsion and supports research on BURs.

\section{CONCLUSIONS AND PERSPECTIVES ON FUTURE DIRECTIONS}
\begin{summary}[Conclusions]
The control of marine robots is undergoing a transformative shift, fueled by rapid advances in data-driven intelligence. This review has explored how data-driven algorithms are reshaping conventional control paradigms for both individual and cooperative marine robotic systems. By surveying recent methods and open-source tools, this work highlights the growing potential of data-driven techniques to enhance the robustness, adaptability, and efficiency of control for both single and multiple marine robotic systems operating in challenging environments.
While recent progress is promising, significant challenges remain, particularly in data efficiency, model interpretability, and onboard deployment. Overcoming these barriers will require interdisciplinary collaboration and a sustained focus on merging theoretical rigor with practical insights. 
In summary, data-driven intelligence is not a replacement for classical and modern control methods, but a powerful complement that enhances their capability. By embracing hybrid approaches, reinforcing theoretical foundations, and emphasizing real-world validation, the community can accelerate progress toward marine robots that operate safely, efficiently, and autonomously in the vast and dynamic marine environments.
\end{summary}

\begin{issues}[FUTURE ISSUES]
\begin{enumerate}
\item \textbf{Integration of traditional control methods with RL}. Traditional methods remain a cornerstone in the control of marine robots due to their reliability and theoretical guarantees. However, they often struggle to handle unmodeled dynamics, uncertainties, nonlinearities, and environmental disturbances. In contrast, RL offers adaptive, data-driven policy optimization but suffers from sample inefficiency, unstable training, and limited generalization due to the sim-to-real gap. A promising solution is the integration of traditional control methods with RL, wherein RL complements rather than replaces conventional schemes.
Residual RL~\cite{johannink2019residual,liu2022deep} exemplifies this hybrid framework: a baseline controller, such as a PID or $H_\infty$ controller, ensures basic stability and safety, while a residual policy learned via RL incrementally improves performance. This design leverages the stability of traditional control methods and the adaptability of data-driven algorithms (e.g., RL), thereby ensuring baseline safety while enhancing system resilience to model uncertainties and environmental disturbances.

\item \textbf{Physical law-constrained data-driven control schemes.} 
Data-driven control methods, such as NN-based approaches~\cite{zhang2022neural}, offer greater adaptability but often lack physical consistency and interpretability, limiting their use in safety-critical systems. PINNs address this limitation by embedding physical laws, such as governing equations and motion constraints, into the learning process. This integration improves model generalization, ensures physical plausibility, and enhances robustness.
PINNs have also demonstrated strong potential in RL-based control methods, where challenges such as data inefficiency and the sim-to-real gap remain significant~\cite{li2024motion,rodwell2023physics}. By incorporating domain knowledge into data-driven models, PINN-based control methods can increase trustworthiness, improve data efficiency, and reduce the discrepancy between simulation and real-world deployment compared to traditional black-box algorithms.

\item \textbf{Large language models (LLMs) empowered control schemes.} LLMs, such as GPT~\cite{achiam2023gpt} and DeepSeek~\cite{liu2024deepseek}, have demonstrated strong capabilities in reasoning, language understanding, and sequential decision-making. Integrating LLMs into the control systems of marine robots presents a promising direction toward cognitive intelligence and high-level autonomy~\cite{10489910,yang2023oceanchat}. LLMs can enhance control frameworks in several impactful ways:
\emph{1)} Interpreting natural language mission descriptions and translating them into structured control objectives for high-level task planning;
\emph{2)} Dynamically generating adaptive control behaviors by reasoning over complex, multi-objective missions and real-time environmental feedback;
\emph{3)} Enabling few-shot adaptation and knowledge transfer by generalizing learned behaviors to novel marine scenarios with minimal retraining;
\emph{4)} Facilitating intuitive human–robot interaction through natural language interfaces, thereby reducing operational complexity and enhancing supervisory control;
\emph{5)} Improving RL by providing semantic priors, initializing policies, or abstracting state representations to boost learning efficiency, stability, and safety.

\item \textbf{Large-scale cooperative control of multiple marine robots.} As underwater missions grow in complexity and scale, large-scale cooperative control of marine robots has become increasingly important. Such cooperation facilitates efficient task execution, functional complementarity, and shared information processing, thereby enhancing system efficiency, robustness, and resilience~\cite{open6}. However, effective large-scale cooperative control of marine robots remains a major challenge due to the limitations of underwater communication, specifically, low-bandwidth, high-latency, data packet loss resulting from the absence of reliable electromagnetic propagation, and susceptibility to network attacks caused by potential human interference~\cite{open7}. These constraints impede real-time synchronization and degrade large-scale or swarm-level coordination. Addressing these issues requires the development of both reliable, low-latency underwater communication infrastructures and robust coordination strategies.

\item \textbf{Mutually reinforcing between simulators and physical experiments.} High-fidelity simulators and physical experiments serve complementary roles in the development of marine robot control systems. Simulators offer safe, scalable, and cost-effective platforms for rapid prototyping and testing across diverse scenarios, but often rely on simplified models that may not fully capture real-world complexities such as sensor noise, environmental disturbances, and unmodeled dynamics~\cite{potokar2024holoocean}. In contrast, physical experiments provide essential real-world validation but are limited by safety risks, environmental unpredictability, and reduced repeatability. When effectively integrated, simulations and experiments form a mutually reinforcing loop: experimental data improve simulator fidelity, while simulators facilitate the design, tuning, and validation of control strategies before real-world deployment. This synergy enhances robustness, supports adaptability, and accelerates the transition from theoretical research to practical application.
\end{enumerate}
\end{issues}

\section*{DISCLOSURE STATEMENT}
The authors are not aware of any affiliations, memberships, funding, or financial holdings that might be perceived as affecting the objectivity of this review.

\section*{ACKNOWLEDGMENTS}
The work described in this paper was partially supported by grants AoE/E-601/24-N, 16203223, and C6029-23G from the Research Grants Council of the Hong Kong Special Administrative Region, China.

\bibliographystyle{ar-style3}
\bibliography{main}

\begin{thebibliography}{150}
\expandafter\ifx\csname natexlab\endcsname\relax\def\natexlab#1{#1}\fi

\bibitem{zhang2015future}
Zhang F, Marani G, Smith RN, Choi HT. 2015.
Future trends in marine robotics [tc spotlight].
\textit{IEEE Robotics \& Automation Magazine} 22(1):14--122

\bibitem{campos2024nautilus}
Campos DF, Gon{\c{c}}alves EP, Campos HJ, Pereira MI, Pinto AM. 2024.
Nautilus: An autonomous surface vehicle with a multilayer software architecture for offshore inspection.
\textit{Journal of Field Robotics} 41(4):966--990

\bibitem{katzschmann2018exploration}
Katzschmann RK, DelPreto J, MacCurdy R, Rus D. 2018.
Exploration of underwater life with an acoustically controlled soft robotic fish.
\textit{Science Robotics} 3(16):eaar3449

\bibitem{wang2023versatile}
Wang T, Joo HJ, Song S, Hu W, Keplinger C, Sitti M. 2023{\natexlab{a}}.
A versatile jellyfish-like robotic platform for effective underwater propulsion and manipulation.
\textit{Science Advances} 9(15):eadg0292

\bibitem{10814088}
Liu Y, Li C, Li J, Lin Z, Meng W, Zhang F. 2025{\natexlab{a}}.
Wukong: Design, modeling and control of a compact flexible hybrid aerial-aquatic vehicle.
\textit{IEEE Robotics and Automation Letters} 10(2):1417--1424

\bibitem{leonard2010coordinated}
Leonard NE, Paley DA, Davis RE, Fratantoni DM, Lekien F, Zhang F. 2010.
Coordinated control of an underwater glider fleet in an adaptive ocean sampling field experiment in monterey bay.
\textit{Journal of Field Robotics} 27(6):718--740

\bibitem{paull2018probabilistic}
Paull L, Seto M, Leonard JJ, Li H. 2018.
Probabilistic cooperative mobile robot area coverage and its application to autonomous seabed mapping.
\textit{The International Journal of Robotics Research} 37(1):21--45

\bibitem{wang2017cooperative}
Wang Y, Garcia E, Casbeer D, Zhang F. 2017.
Cooperative control of multi-agent systems: Theory and applications

\bibitem{chang2017motion}
Chang D, Wu W, Edwards CR, Zhang F. 2017.
Motion tomography: Mapping flow fields using autonomous underwater vehicles.
\textit{The International Journal of Robotics Research} 36(3):320--336

\bibitem{Heshmati-JOE-2020}
Heshmati-Alamdari S, Bechlioulis CP, Karras GC, Kyriakopoulos KJ. 2020.
Cooperative impedance control for multiple underwater vehicle manipulator systems under lean communication.
\textit{IEEE Journal of Oceanic Engineering} 46(2):447--465

\bibitem{degorre2023survey}
Degorre L, Delaleau E, Chocron O. 2023.
A survey on model-based control and guidance principles for autonomous marine vehicles.
\textit{Journal of Marine Science and Engineering} 11(2):430

\bibitem{he2020review}
He Y, Wang DB, Ali ZA. 2020.
A review of different designs and control models of remotely operated underwater vehicle.
\textit{Measurement and Control} 53(9-10):1561--1570

\bibitem{wang2019state}
Wang L, Wu Q, Liu J, Li S, Negenborn RR. 2019{\natexlab{a}}.
State-of-the-art research on motion control of maritime autonomous surface ships.
\textit{Journal of Marine Science and Engineering} 7(12):438

\bibitem{er2023intelligent}
Er MJ, Ma C, Liu T, Gong H. 2023.
Intelligent motion control of unmanned surface vehicles: A critical review.
\textit{Ocean Engineering} 280:114562

\bibitem{wang2022development}
Wang J, Wu Z, Dong H, Tan M, Yu J. 2022{\natexlab{a}}.
Development and control of underwater gliding robots: A review.
\textit{IEEE/CAA Journal of Automatica Sinica} 9(9):1543--1560

\bibitem{wang2020development}
Wang R, Wang S, Wang Y, Cheng L, Tan M. 2020.
Development and motion control of biomimetic underwater robots: A survey.
\textit{IEEE Transactions on Systems, Man, and Cybernetics: Systems} 52(2):833--844

\bibitem{sun2022recent}
Sun B, Li W, Wang Z, Zhu Y, He Q, et~al. 2022.
Recent progress in modeling and control of bio-inspired fish robots.
\textit{Journal of Marine Science and Engineering} 10(6):773

\bibitem{xiang2018survey}
Xiang X, Yu C, Lapierre L, Zhang J, Zhang Q. 2018.
Survey on fuzzy-logic-based guidance and control of marine surface vehicles and underwater vehicles.
\textit{International Journal of Fuzzy Systems} 20:572--586

\bibitem{wei2022mpc}
Wei H, Shi Y. 2022.
Mpc-based motion planning and control enables smarter and safer autonomous marine vehicles: Perspectives and a tutorial survey.
\textit{IEEE/CAA Journal of Automatica Sinica} 10(1):8--24

\bibitem{hassani2018data}
Hassani V, Pascoal AM, Onstein TF. 2018.
Data-driven control in marine systems.
\textit{Annual Reviews in Control} 46:343--349

\bibitem{tijjani2022survey}
Tijjani AS, Chemori A, Creuze V. 2022.
A survey on tracking control of unmanned underwater vehicles: Experiments-based approach.
\textit{Annual Reviews in Control} 54:125--147

\bibitem{das2016cooperative}
Das B, Subudhi B, Pati BB. 2016.
Cooperative formation control of autonomous underwater vehicles: An overview.
\textit{International Journal of Automation and computing} 13:199--225

\bibitem{peng2020overview}
Peng Z, Wang J, Wang D, Han QL. 2020{\natexlab{a}}.
An overview of recent advances in coordinated control of multiple autonomous surface vehicles.
\textit{IEEE Transactions on Industrial Informatics} 17(2):732--745

\bibitem{yang2021survey}
Yang Y, Xiao Y, Li T. 2021.
A survey of autonomous underwater vehicle formation: Performance, formation control, and communication capability.
\textit{IEEE Communications Surveys \& Tutorials} 23(2):815--841

\bibitem{jiang2022neural}
Jiang T, Yan Y, Wu D, Yu S, Li T. 2022{\natexlab{a}}.
Neural network based adaptive sliding mode tracking control of autonomous surface vehicles with input quantization and saturation.
\textit{Ocean Engineering} 265:112505

\bibitem{8714020}
Chen L, Cui R, Yang C, Yan W. 2020.
Adaptive neural network control of underactuated surface vessels with guaranteed transient performance: Theory and experimental results.
\textit{IEEE Transactions on Industrial Electronics} 67(5):4024--4035

\bibitem{zhang2022neural}
Zhang JX, Yang T, Chai T. 2022.
Neural network control of underactuated surface vehicles with prescribed trajectory tracking performance.
\textit{IEEE Transactions on Neural Networks and Learning Systems}

\bibitem{li2024gaussian}
Li F, Li H, Wu C. 2024.
Gaussian process-based learning model predictive control with application to usv.
\textit{IEEE Transactions on Industrial Electronics}

\bibitem{lima2020sliding}
Lima GS, Trimpe S, Bessa WM. 2020.
Sliding mode control with gaussian process regression for underwater robots.
\textit{Journal of Intelligent \& Robotic Systems} 99(3):487--498

\bibitem{dang2025incremental}
Dang Y, Huang Y, Shen X, Zhu D, Chu Z. 2025.
Incremental sparse gaussian process-based model predictive control for trajectory tracking of unmanned underwater vehicles.
\textit{IEEE Robotics and Automation Letters}

\bibitem{amer2025empowering}
Amer A, Mehndiratta M, Brodskiy Y, Kayacan E. 2025.
Empowering autonomous underwater vehicles using learning-based model predictive control with dynamic forgetting gaussian processes.
\textit{IEEE Transactions on Control Systems Technology}

\bibitem{cui2022filtered}
Cui Y, Peng L, Li H. 2022.
Filtered probabilistic model predictive control-based reinforcement learning for unmanned surface vehicles.
\textit{IEEE Transactions on Industrial Informatics} 18(10):6950--6961

\bibitem{rahmani2024enhanced}
Rahmani M, Redkar S. 2024.
Enhanced koopman operator-based robust data-driven control for 3 degree of freedom autonomous underwater vehicles: A novel approach.
\textit{Ocean Engineering} 307:118227

\bibitem{li2024c3d}
Li J, Park H, Hao W, Xin L, Chavez-Galaviz J, et~al. 2024{\natexlab{a}}.
C3d: Cascade control with change point detection and deep koopman learning for autonomous surface vehicles.
\textit{arXiv preprint arXiv:2403.05972}

\bibitem{mamakoukas2021derivative}
Mamakoukas G, Castano ML, Tan X, Murphey TD. 2021.
Derivative-based koopman operators for real-time control of robotic systems.
\textit{IEEE Transactions on Robotics} 37(6):2173--2192

\bibitem{cong2025koopman}
Cong B, Liu Z. 2025.
\textit{Koopman Operator and Model Predictive Control Integration for Autonomous Underwater Vehicle Speed Regulation}.
In \textit{2025 Australian \& New Zealand Control Conference (ANZCC)}, pp.  81--86. IEEE

\bibitem{raissi2019physics}
Raissi M, Perdikaris P, Karniadakis GE. 2019.
Physics-informed neural networks: A deep learning framework for solving forward and inverse problems involving nonlinear partial differential equations.
\textit{Journal of Computational physics} 378:686--707

\bibitem{liu2024research}
Liu T, Zhao J, Huang J, Li Z, Xu L, Zhao B. 2024{\natexlab{a}}.
Research on model predictive control of autonomous underwater vehicle based on physics informed neural network modeling.
\textit{Ocean Engineering} 304:117844

\bibitem{majumder2024safe}
Majumder R, Makam R, Mane P, KS B, Sundaram S. 2024.
\textit{Safe navigation of autonomous underwater vehicles using physics-informed neural networks}.
In \textit{OCEANS 2024-Singapore}, pp.  1--6. IEEE

\bibitem{algarin2025intelligent}
Algar{\'\i}n-Pinto JA, Garza-Casta{\~n}{\'o}n LE, Vargas-Mart{\'\i}nez A, Minchala-{\'A}vila LI, Payeur P. 2025.
Intelligent motion control to enhance the swimming performance of a biomimetic underwater vehicle using reinforcement learning approach.
\textit{IEEE Access}

\bibitem{higo2023development}
Higo Y, Sakano M, Nobe H, Hashimoto H. 2023.
Development of trajectory-tracking maneuvering system for automatic berthing/unberthing based on double deep q-network and experimental validation with an actual large ferry.
\textit{Ocean Engineering} 287:115750

\bibitem{wu2020autonomous}
Wu X, Chen H, Chen C, Zhong M, Xie S, et~al. 2020.
The autonomous navigation and obstacle avoidance for usvs with anoa deep reinforcement learning method.
\textit{Knowledge-Based Systems} 196:105201

\bibitem{chen2024deep}
Chen G, Zhao Z, Lu Y, Yang C, Hu H. 2024.
Deep reinforcement learning-based pitch attitude control of a beaver-like underwater robot.
\textit{Ocean Engineering} 307:118163

\bibitem{wang2022deep}
Wang X, Wang S, Liang X, Zhao D, Huang J, et~al. 2022{\natexlab{b}}.
Deep reinforcement learning: A survey.
\textit{IEEE Transactions on Neural Networks and Learning Systems} 35(4):5064--5078

\bibitem{overeng2021dynamic}
{\O}vereng SS, Nguyen DT, Hamre G. 2021.
Dynamic positioning using deep reinforcement learning.
\textit{Ocean Engineering} 235:109433

\bibitem{hasankhani2023integrated}
Hasankhani A, Tang Y, VanZwieten J. 2023.
Integrated path planning and control through proximal policy optimization for a marine current turbine.
\textit{Applied Ocean Research} 137:103591

\bibitem{wu2023deep}
Wu C, Yu W, Li G, Liao W. 2023.
Deep reinforcement learning with dynamic window approach based collision avoidance path planning for maritime autonomous surface ships.
\textit{Ocean Engineering} 284:115208

\bibitem{huang2024learning}
Huang H, Jiang T, Zhang Z, Sun Y, Qin H, et~al. 2024.
Learning strategies for underwater robot autonomous manipulation control.
\textit{Journal of the Franklin Institute} 361(7):106773

\bibitem{hao2024target}
Hao Y, Wang Q, Shen K. 2024.
\textit{Target Tracking Control of Underactuated Unmanned Boats Using an Improved PPO Algorithm}.
In \textit{International Conference on Guidance, Navigation and Control}, pp.  330--340. Springer

\bibitem{zhang2023auv}
Zhang T, Miao X, Li Y, Jia L, Wei Z, et~al. 2023.
Auv 3d docking control using deep reinforcement learning.
\textit{Ocean engineering} 283:115021

\bibitem{chu2025adaptive}
Chu S, Lin M, Li D, Lin R, Xiao S. 2025.
Adaptive reward shaping based reinforcement learning for docking control of autonomous underwater vehicles.
\textit{Ocean Engineering} 318:120139

\bibitem{wang2021data}
Wang N, Gao Y, Zhang X. 2021.
Data-driven performance-prescribed reinforcement learning control of an unmanned surface vehicle.
\textit{IEEE Transactions on Neural Networks and Learning Systems} 32(12):5456--5467

\bibitem{deng2021event}
Deng Y, Liu T, Zhao D. 2021.
Event-triggered output-feedback adaptive tracking control of autonomous underwater vehicles using reinforcement learning.
\textit{Applied Ocean Research} 113:102676

\bibitem{wang2024reinforcement}
Wang Y, Gao J. 2024.
Reinforcement-learning-based visual servoing of underwater vehicle dual-manipulator system.
\textit{Journal of Marine Science and Engineering} 12(6):940

\bibitem{wang2024learning}
Wang Y, Chu H, Ma R, Bai X, Cheng L, et~al. 2024.
Learning-based discontinuous path following control for a biomimetic underwater vehicle.
\textit{Research} 7:0299

\bibitem{ma2023position}
Ma R, Wang Y, Tang C, Wang S, Wang R. 2023{\natexlab{a}}.
Position and attitude tracking control of a biomimetic underwater vehicle via deep reinforcement learning.
\textit{IEEE/ASME Transactions on Mechatronics} 28(5):2810--2819

\bibitem{wang2025expert}
Wang Y, Hou Y, Lai Z, Cao L, Hong W, Wu D. 2025{\natexlab{a}}.
An expert-demonstrated soft actor--critic based adaptive trajectory tracking control of autonomous underwater vehicle with long short-term memory.
\textit{Ocean Engineering} 321:120405

\bibitem{yuan2023deep}
Yuan W, Rui X. 2023.
Deep reinforcement learning-based controller for dynamic positioning of an unmanned surface vehicle.
\textit{Computers and Electrical Engineering} 110:108858

\bibitem{dong2025end}
Dong N, Liu S, Ip AW, Yung KL, Gao Z, et~al. 2025.
End-to-end autonomous underwater vehicle path following control method based on improved soft actor-critic for deep space exploration.
\textit{Journal of Industrial Information Integration} :100792

\bibitem{ma2023sample}
Ma R, Wang Y, Wang S, Cheng L, Wang R, Tan M. 2023{\natexlab{b}}.
Sample-observed soft actor-critic learning for path following of a biomimetic underwater vehicle.
\textit{IEEE Transactions on Automation Science and Engineering}

\bibitem{sun2020auv}
Sun Y, Ran X, Zhang G, Wang X, Xu H. 2020.
Auv path following controlled by modified deep deterministic policy gradient.
\textit{Ocean Engineering} 210:107360

\bibitem{9351698}
Wang Y, Tang C, Wang S, Cheng L, Wang R, et~al. 2022{\natexlab{c}}.
Target tracking control of a biomimetic underwater vehicle through deep reinforcement learning.
\textit{IEEE Transactions on Neural Networks and Learning Systems} 33(8):3741--3752

\bibitem{zhang2022auv}
Zhang C, Cheng P, Du B, Dong B, Zhang W. 2022{\natexlab{a}}.
Auv path tracking with real-time obstacle avoidance via reinforcement learning under adaptive constraints.
\textit{Ocean Engineering} 256:111453

\bibitem{masmitja2023dynamic}
Masmitja I, Martin M, O’Reilly T, Kieft B, Palomeras N, et~al. 2023.
Dynamic robotic tracking of underwater targets using reinforcement learning.
\textit{Science robotics} 8(80):eade7811

\bibitem{du2022safe}
Du B, Lin B, Zhang C, Dong B, Zhang W. 2022{\natexlab{a}}.
Safe deep reinforcement learning-based adaptive control for usv interception mission.
\textit{Ocean Engineering} 246:110477

\bibitem{zhu2022autonomous}
Zhu P, Liu S, Jiang T, Liu Y, Zhuang X, Zhang Z. 2022.
Autonomous reinforcement control of visual underwater vehicles: Real-time experiments using computer vision.
\textit{IEEE Transactions on Vehicular Technology} 71(8):8237--8250

\bibitem{grando2021deep}
Grando RB, de~Jesus JC, Kich VA, Kolling AH, Bortoluzzi NP, et~al. 2021.
\textit{Deep reinforcement learning for mapless navigation of a hybrid aerial underwater vehicle with medium transition}.
In \textit{2021 IEEE International Conference on Robotics and Automation (ICRA)}, pp.  1088--1094. IEEE

\bibitem{hadi2022deep}
Hadi B, Khosravi A, Sarhadi P. 2022.
Deep reinforcement learning for adaptive path planning and control of an autonomous underwater vehicle.
\textit{Applied Ocean Research} 129:103326

\bibitem{fan2024path}
Fan Y, Dong H, Zhao X, Denissenko P. 2024.
Path-following control of unmanned underwater vehicle based on an improved td3 deep reinforcement learning.
\textit{IEEE Transactions on Control Systems Technology}

\bibitem{zhou2025adaptive}
Zhou Y, Gong C, Chen K. 2025.
Adaptive control scheme for usv trajectory-tracking under complex environmental disturbances via deep reinforcement learning.
\textit{IEEE Internet of Things Journal}

\bibitem{sivaraj2023performance}
Sivaraj S, Dubey A, Rajendran S. 2023.
On the performance of different deep reinforcement learning based controllers for the path-following of a ship.
\textit{Ocean Engineering} 286:115607

\bibitem{manderson_vision-based_2020}
Manderson T, Higuera JCG, Wapnick S, Tremblay JF, Shkurti F, et~al. 2020.
\textit{{Vision-Based Goal-Conditioned Policies for Underwater Navigation in the Presence of Obstacles}}.
In \textit{Proceedings of Robotics: Science and Systems}. Corvalis, Oregon, USA

\bibitem{lin2024uivnav}
Lin X, Karapetyan N, Joshi K, Liu T, Chopra N, et~al. 2024.
\textit{Uivnav: Underwater information-driven vision-based navigation via imitation learning}.
In \textit{2024 IEEE International Conference on Robotics and Automation (ICRA)}, pp.  5250--5256. IEEE

\bibitem{liu2024self}
Liu R, Ha H, Hou M, Song S, Vondrick C. 2024{\natexlab{b}}.
Self-improving autonomous underwater manipulation.
\textit{arXiv preprint arXiv:2410.18969}

\bibitem{zhang2022leveraging}
Zhang T, Yue L, Wang C, Sun J, Zhang S, et~al. 2022{\natexlab{b}}.
Leveraging imitation learning on pose regulation problem of a robotic fish.
\textit{IEEE Transactions on Neural Networks and Learning Systems} 35(3):4232--4245

\bibitem{goodfellow2020generative}
Goodfellow I, Pouget-Abadie J, Mirza M, Xu B, Warde-Farley D, et~al. 2020.
Generative adversarial networks.
\textit{Communications of the ACM} 63(11):139--144

\bibitem{chaysri2023unmanned}
Chaysri P, Spatharis C, Blekas K, Vlachos K. 2023.
Unmanned surface vehicle navigation through generative adversarial imitation learning.
\textit{Ocean Engineering} 282:114989

\bibitem{jiang2022generative}
Jiang D, Huang J, Fang Z, Cheng C, Sha Q, et~al. 2022{\natexlab{b}}.
Generative adversarial interactive imitation learning for path following of autonomous underwater vehicle.
\textit{Ocean Engineering} 260:111971

\bibitem{chu2020motion}
Chu Z, Sun B, Zhu D, Zhang M, Luo C. 2020.
Motion control of unmanned underwater vehicles via deep imitation reinforcement learning algorithm.
\textit{IET Intelligent Transport Systems} 14(7):764--774

\bibitem{yan2022real}
Yan S, Wu Z, Wang J, Huang Y, Tan M, Yu J. 2022.
Real-world learning control for autonomous exploration of a biomimetic robotic shark.
\textit{IEEE Transactions on Industrial Electronics} 70(4):3966--3974

\bibitem{martinsen2022reinforcement}
Martinsen AB, Lekkas AM, Gros S. 2022.
Reinforcement learning-based nmpc for tracking control of asvs: Theory and experiments.
\textit{Control Engineering Practice} 120:105024

\bibitem{zhang2024deep}
Zhang Z, Pan X, Chen T, Jiang D, Fang Z, Li G. 2024{\natexlab{a}}.
\textit{Deep Reinforcement Learning with Model Predictive Control for Path Following of Autonomous Underwater Vehicle}.
In \textit{2024 43rd Chinese Control Conference (CCC)}, pp.  2516--2523. IEEE

\bibitem{ma2023neural}
Ma D, Chen X, Ma W, Zheng H, Qu F. 2023{\natexlab{c}}.
Neural network model-based reinforcement learning control for auv 3-d path following.
\textit{IEEE Transactions on Intelligent Vehicles} 9(1):893--904

\bibitem{cui2021autonomous}
Cui Y, Osaki S, Matsubara T. 2021.
Autonomous boat driving system using sample-efficient model predictive control-based reinforcement learning approach.
\textit{Journal of Field Robotics} 38(3):331--354

\bibitem{pan2022learning}
Pan J, Zhang P, Wang J, Liu M, Yu J. 2022.
Learning for depth control of a robotic penguin: A data-driven model predictive control approach.
\textit{IEEE Transactions on Industrial Electronics} 70(11):11422--11432

\bibitem{8734349}
Wang D, Shen Y, Sha Q, Li G, Kong X, et~al. 2019{\natexlab{b}}.
\textit{Adaptive DDPG Design-Based Sliding-Mode Control for Autonomous Underwater Vehicles at Different Speeds}.
In \textit{2019 IEEE Underwater Technology (UT)}, pp.  1--5

\bibitem{zhu2025sliding}
Zhu S, Zhang G, Wang Q, Li Z. 2025.
Sliding mode control for variable-speed trajectory tracking of underactuated vessels with td3 algorithm optimization.
\textit{Journal of Marine Science and Engineering} 13(1):99

\bibitem{wang2022sliding}
Wang D, Shen Y, Wan J, Sha Q, Li G, et~al. 2022{\natexlab{d}}.
Sliding mode heading control for auv based on continuous hybrid model-free and model-based reinforcement learning.
\textit{Applied Ocean Research} 118:102960

\bibitem{Peng-TC-2020}
Peng Z, Wang D, Li T, Han M. 2020{\natexlab{b}}.
Output-feedback cooperative formation maneuvering of autonomous surface vehicles with connectivity preservation and collision avoidance.
\textit{IEEE Transactions on Cybernetics} 50(6):2527--2535

\bibitem{Jiang-OE-2024-Adaptive}
Jiang Y, Liu Z, Chen F. 2024.
Adaptive output-constrained finite-time formation control for multiple unmanned surface vessels with directed communication topology.
\textit{Ocean Engineering} 292:116552

\bibitem{Guo-AOR-2024-adaptive}
Guo Q, Zhang X, Ma D. 2024.
Adaptive fixed-time path following cooperative control for autonomous surface vehicles based on path-dependent constraints.
\textit{Applied Ocean Research} 142:103826

\bibitem{Ma-JAS-2023}
Ma L, Wang YL, Han QL. 2023.
Cooperative target tracking of multiple autonomous surface vehicles under switching interaction topologies.
\textit{IEEE/CAA Journal of Automatica Sinica} 10(3):673--684

\bibitem{Sahu-TFS-2018}
Sahu BK, Subudhi B. 2018.
Flocking control of multiple auvs based on fuzzy potential functions.
\textit{IEEE Transactions on Fuzzy Systems} 26(5):2539--2551

\bibitem{Peng-TSMCS-2023}
Peng Z, Jiang Y, Liu L, Shi Y. 2023.
Path-guided model-free flocking control of unmanned surface vehicles based on concurrent learning extended state observers.
\textit{IEEE Transactions on Systems, Man, and Cybernetics: Systems} 53(8):4729--4739

\bibitem{Peng-TC-2021}
Peng Z, Liu L, Wang J. 2021.
Output-feedback flocking control of multiple autonomous surface vehicles based on data-driven adaptive extended state observers.
\textit{IEEE Transactions on Cybernetics} 51(9):4611--4622

\bibitem{Zhang-OE-2025}
Zhang C, Zeng R, Lin B, Zhang Y, Xie W, Zhang W. 2025.
Multi-usv cooperative target encirclement through learning-based distributed transferable policy and experimental validation.
\textit{Ocean Engineering} 318:120124

\bibitem{Jiang-TFS-2024}
Jiang Y, Peng Z, Liu L, Wang D, Zhang F. 2024{\natexlab{a}}.
Safety-critical cooperative target enclosing control of autonomous surface vehicles based on finite-time fuzzy predictors and input-to-state safe high-order control barrier functions.
\textit{IEEE Transactions on Fuzzy Systems} 32(3):816--830

\bibitem{Yan-ASME-2025}
Yan Z, Zheng H, Jiang Z, Xu W. 2025.
Distributed control of unmanned marine vehicles for target circumnavigation in communication-denied environments.
\textit{IEEE/ASME Transactions on Mechatronics} 30(1):345--356

\bibitem{Luo-OE-2024}
Luo J, Su Y. 2024.
Path planning for multi-usv target coverage in complex environments.
\textit{Ocean Engineering} 312:119090

\bibitem{Jiao-OE-2025}
Jiao S, Liu L, Peng Z, Li T, Zhang W. 2025.
Collision-free dynamic coverage of autonomous surface vehicles with anisotropic sensing.
\textit{Ocean Engineering} 319:120163

\bibitem{Liu-TVT-2025}
Liu L, Jiao S, Han B, Li T, Peng Z. 2025{\natexlab{b}}.
Swarm-based dynamic coverage of multi-asv systems in the presence of measurement noises.
\textit{IEEE Transactions on Vehicular Technology} :1--15

\bibitem{Li-TSMCS-2024}
Li F, Yin M, Wang T, Huang T, Yang C, Gui W. 2024{\natexlab{b}}.
Distributed pursuit-evasion game of limited perception {USV} swarm based on multiagent proximal policy optimization.
\textit{IEEE Transactions on Systems, Man, and Cybernetics: Systems}

\bibitem{Meng-TAI-2024}
Meng Y, Liu C, Wang Q, Tan L. 2024.
Cooperative advantage actor-critic reinforcement learning for multi-agent pursuit-evasion games on communication graphs.
\textit{IEEE Transactions on Artificial Intelligence}

\bibitem{Jiang-TIV-2024}
Jiang Y, Li Z. 2024.
Fully distributed target encircling control of autonomous surface vehicles based on noncooperative games.
\textit{IEEE Transactions on Intelligent Vehicles} 9(4):4769--4779

\bibitem{Nantogma-Ele-2023}
Nantogma S, Zhang S, Yu X, An X, Xu Y. 2023.
Multi-usv dynamic navigation and target capture: A guided multi-agent reinforcement learning approach.
\textit{Electronics} 12(7):1523

\bibitem{Kang-TIE-2024}
Kang T, Gu N, Wang D, Liu L, Hu Q, Peng Z. 2024.
Neurodynamics-based attack-defense guidance of autonomous surface vehicles against multiple attackers for domain protection.
\textit{IEEE Transactions on Industrial Electronics} 71(10):12655--12663

\bibitem{Du-OE-2022}
Du B, Lin B, Zhang C, Dong B, Zhang W. 2022{\natexlab{b}}.
Safe deep reinforcement learning-based adaptive control for usv interception mission.
\textit{Ocean Engineering} 246:110477

\bibitem{Feng-ELL-2025}
Feng D, Yang J, Zhang N, Xiao J, Dai S, et~al. 2025.
Study on key technologies for air--water surface collaboration of observation unmanned aircraft vehicle.
\textit{Electronics Letters} 61(1):e70164

\bibitem{Jiang-SSCAE-2024}
Jiang B, Wen G, Zhou J, Zheng D. 2024{\natexlab{b}}.
Cross-domain cooperative technology of intelligent unmanned swarm systems: Current status and prospects.
\textit{Strategic Study of Chinese Academy of Engineering} 26(1):117--126

\bibitem{Cao-TIV-2024}
Cao X, Liu W, Ren L. 2024.
Underwater target capture based on heterogeneous unmanned system collaboration.
\textit{IEEE Transactions on Intelligent Vehicles}

\bibitem{Liu-JOE-2023}
Liu Y, Chen C, Qu D, Zhong Y, Pu H, et~al. 2023.
Multi-{USV} system antidisturbance cooperative searching based on the reinforcement learning method.
\textit{IEEE Journal of Oceanic Engineering} 48(4):1019--1047

\bibitem{Wang-OE-2023}
Wang Z, Zhang L, Zhu Z. 2023.
Game-based distributed optimal formation tracking control of underactuated {AUV}s based on reinforcement learning.
\textit{Ocean Engineering} 287:115879

\bibitem{Cao-JAS-2023}
Cao W, Yan J, Yang X, Luo X, Guan X. 2023.
Communication-aware formation control of {AUV}s with model uncertainty and fading channel via integral reinforcement learning.
\textit{IEEE/CAA Journal of Automatica Sinica} 10(1):159--176

\bibitem{Wang-TNSE-2023}
Wang P, Yu C, Lv M, Cao J. 2023{\natexlab{b}}.
Adaptive fixed-time optimal formation control for uncertain nonlinear multiagent systems using reinforcement learning.
\textit{IEEE Transactions on Network Science and Engineering} 11(2):1729--1743

\bibitem{Ding-TIV-2023}
Ding TF, Ge MF, Liu ZW, Wang L, Liu J. 2023.
Reinforcement learning formation tracking of networked autonomous surface vehicles with bounded inputs via cloud-supported communication.
\textit{IEEE Transactions on Intelligent Vehicles} 9(1):469--480

\bibitem{Ding-TFS-2024}
Ding TF, Zhang HY, Ge MF, Liu ZW. 2024.
Predefined-time fuzzy reinforcement learning control for secure surrounding formation of nmsvs with dos attacks.
\textit{IEEE Transactions on Fuzzy Systems}

\bibitem{Zhang-ITJ-2024}
Zhang J, Ren J, Cui Y, Fu D, Cong J. 2024{\natexlab{b}}.
Multi-usv task planning method based on improved deep reinforcement learning.
\textit{IEEE Internet of Things Journal} 11(10):18549--18567

\bibitem{Gan-TASE-2023}
Gan W, Qu X, Song D, Yao P. 2023.
Multi-usv cooperative chasing strategy based on obstacles assistance and deep reinforcement learning.
\textit{IEEE Transactions on Automation Science and Engineering}

\bibitem{Wang-TITS-2025}
Wang Z, Chen P, Chen L, Mou J. 2025{\natexlab{b}}.
Collaborative collision avoidance approach for {USV}s based on multi-agent deep reinforcement learning.
\textit{IEEE Transactions on Intelligent Transportation Systems}

\bibitem{Xu-OE-2021}
Xu J, Huang F, Wu D, Cui Y, Yan Z, Zhang K. 2021.
Deep reinforcement learning based multi-auvs cooperative decision-making for attack--defense confrontation missions.
\textit{Ocean Engineering} 239:109794

\bibitem{Peng-TCST-2018}
Peng Z, Wang J, Wang D. 2017.
Distributed maneuvering of autonomous surface vehicles based on neurodynamic optimization and fuzzy approximation.
\textit{IEEE Transactions on Control Systems Technology} 26(3):1083--1090

\bibitem{Peng-OE-2022}
Peng Z, Jiang Y, Liu L, Wang D. 2022.
Distributed optimization for coordinated dynamic positioning of multiple surface vessels based on asymptotically stable esos.
\textit{Ocean Engineering} 246:110507

\bibitem{Lyu-TCYB-2024}
Lyu G, Peng Z, Wang J. 2024.
Safety-critical receding-horizon planning and formation control of autonomous surface vehicles via collaborative neurodynamic optimization.
\textit{IEEE Transactions on Cybernetics} 54(12):7236--7247

\bibitem{Sun-Ne-2022}
Sun Y, Du Y, Qin H. 2022.
Distributed adaptive neural network constraint containment control for the benthic autonomous underwater vehicles.
\textit{Neurocomputing} 484:89--98

\bibitem{Fan-OE-2025}
Fan Y, Li Z, Li J, Ma G, Bu H. 2025.
Fixed-time event-triggered distributed formation control for underactuated usvs considering actuator saturation.
\textit{Ocean Engineering} 316:119829

\bibitem{Liu-TIE-2024}
Liu L, Zhang J, Sun R, Peng Z, Wang D, Shi Y. 2024{\natexlab{c}}.
Cloud-based self-triggered cooperative path following of underactuated usvs with multimodel extended state observers.
\textit{IEEE Transactions on Industrial Electronics}

\bibitem{Wang-TIE-2023}
Wang H, Tan H, Peng Z. 2023.
Quantized communications in containment maneuvering for output constrained marine surface vehicles: Theory and experiment.
\textit{IEEE Transactions on Industrial Electronics} 71(1):880--889

\bibitem{Suryendu-TCSII-2020}
Suryendu C, Subudhi B. 2020.
Formation control of multiple autonomous underwater vehicles under communication delays.
\textit{IEEE Transactions on Circuits and Systems II: Express Briefs} 67(12):3182--3186

\bibitem{Yan-TCST-2019}
Yan J, Gao J, Yang X, Luo X, Guan X. 2019.
Position tracking control of remotely operated underwater vehicles with communication delay.
\textit{IEEE Transactions on Control Systems Technology} 28(6):2506--2514

\bibitem{Jiang-OE-2024}
Jiang Y, Wang L, Sun J, Yu H. 2024{\natexlab{c}}.
Attack-resistant distributed formation control for multiple unmanned surface vessels subject to output constraints.
\textit{Ocean Engineering} 314:119712

\bibitem{Gao-TIV-2022}
Gao S, Peng Z, Liu L, Wang D, Han QL. 2022.
Fixed-time resilient edge-triggered estimation and control of surface vehicles for cooperative target tracking under attacks.
\textit{IEEE Transactions on Intelligent Vehicles} 8(1):547--556

\bibitem{Gu-ISA-2020}
Gu N, Wang D, Peng Z, Liu L. 2020.
Adaptive bounded neural network control for coordinated path-following of networked underactuated autonomous surface vehicles under time-varying state-dependent cyber-attack.
\textit{ISA Transactions} 104:212--221

\bibitem{Jiang-JSYST-2022}
Jiang X, Xia G. 2022.
Nonfragile formation control of leaderless unmanned surface vehicles with memory sampling data and packet loss.
\textit{IEEE Systems Journal} 17(2):3026--3035

\bibitem{Zhou-TASE-2025}
Zhou X, Huang B, Zhou B, Zhu C, Qin H, Miao J. 2025.
{Affine formation maneuver control for NUSVs: An anti-competing interaction solution with random packet losses}.
\textit{IEEE Transactions on Automation Science and Engineering} 22:5916--5932

\bibitem{wu2013glider}
Wu W, Chang D, Zhang F. 2013.
\textit{Glider CT: Reconstructing flow fields from predicted motion of underwater gliders}.
In \textit{Proceedings of the 8th International Conference on Underwater Networks \& Systems}, pp.  1--8

\bibitem{chang2016glider}
Chang D, Wu W, Zhang F. 2016.
\textit{Glider CT: Analysis and experimental validation}.
In \textit{Distributed Autonomous Robotic Systems: The 12th International Symposium}, pp.  285--298. Springer

\bibitem{chang2016distributed}
Chang D, Zhang F. 2016.
\textit{Distributed motion tomography for time-varying flow fields}.
In \textit{OCEANS 2016-Shanghai}, pp.  1--7. IEEE

\bibitem{chang2019distributed}
Chang D, Zhang F, Sun J. 2019.
\textit{Distributed motion tomography for reconstruction of flow fields}.
In \textit{2019 international conference on robotics and automation (ICRA)}, pp.  8048--8054. IEEE

\bibitem{zuo2023bio}
Zuo W, Zhang F, Chen Z. 2023.
Bio-inspired robotic fish enabled motion tomography.
\textit{International Journal of Intelligent Robotics and Applications} 7(3):474--484

\bibitem{johannink2019residual}
Johannink T, Bahl S, Nair A, Luo J, Kumar A, et~al. 2019.
\textit{Residual reinforcement learning for robot control}.
In \textit{2019 international conference on robotics and automation (ICRA)}, pp.  6023--6029. IEEE

\bibitem{liu2022deep}
Liu YT, Price E, Black MJ, Ahmad A. 2022.
\textit{Deep residual reinforcement learning based autonomous blimp control}.
In \textit{2022 IEEE/RSJ International Conference on Intelligent Robots and Systems (IROS)}, pp.  12566--12573. IEEE

\bibitem{li2024motion}
Li X, Geng L, Liu K, Zhao Y, Du W. 2024{\natexlab{c}}.
Motion control of autonomous underwater vehicle based on physics-informed offline reinforcement learning.
\textit{Ocean Engineering} 313:119432

\bibitem{rodwell2023physics}
Rodwell C, Tallapragada P. 2023.
Physics-informed reinforcement learning for motion control of a fish-like swimming robot.
\textit{Scientific Reports} 13(1):10754

\bibitem{achiam2023gpt}
Achiam J, Adler S, Agarwal S, Ahmad L, Akkaya I, et~al. 2023.
Gpt-4 technical report.
\textit{arXiv preprint arXiv:2303.08774}

\bibitem{liu2024deepseek}
Liu A, Feng B, Xue B, Wang B, Wu B, et~al. 2024{\natexlab{d}}.
Deepseek-v3 technical report.
\textit{arXiv preprint arXiv:2412.19437}

\bibitem{10489910}
Chen W, Li G, Li M, Wang W, Li P, et~al. 2025.
Llm-enabled incremental learning framework for hand exoskeleton control.
\textit{IEEE Transactions on Automation Science and Engineering} 22:2617--2626

\bibitem{yang2023oceanchat}
Yang R, Hou M, Wang J, Zhang F. 2023.
Oceanchat: Piloting autonomous underwater vehicles in natural language.
\textit{arXiv preprint arXiv:2309.16052}

\bibitem{open6}
Xiao-long L, Qiang S, Zhong-hai Y, Ya-li W, Ping-ni L. 2015.
Review on large-scale unmanned system swarm intelligence control method.
\textit{Application Research of Computers/Jisuanji Yingyong Yanjiu} 32(1)

\bibitem{open7}
Huang Z, Yuan L, Cai W. 2022.
Research of key technology in marine unmanned swarm commnunication network.
\textit{Application Research of Computers} 44(14):127--132

\bibitem{potokar2024holoocean}
Potokar E, Lay K, Norman K, Benham D, Ashford S, et~al. 2024.
Holoocean: A full-featured marine robotics simulator for perception and autonomy.
\textit{IEEE Journal of Oceanic Engineering}

\end{thebibliography}
\end{document}